\documentclass[runningheads]{llncs}

\newcommand{\supplementarysection}{%
  \let\oldthefigure\thefigure
  \renewcommand{\thefigure}{S\oldthefigure}
  \section{Supplementary}
  \let\oldchapter\section
  \renewcommand{\section}{
    \let\thefigure\oldthefigure
    \let\section\oldchapter
    \oldchapter
  }
}

\usepackage{graphicx}

\usepackage{appendix}
\usepackage{tikz}
\usepackage{comment}
\usepackage{amsmath,amssymb} 
\usepackage{color}
\usepackage{orcidlink}
\usepackage[capitalize]{cleveref}

\usepackage[accsupp]{axessibility}  

\newcommand{\etal}{\textit{et al}.}

\begin{document}
\pagestyle{headings}
\mainmatter
\def\ECCVSubNumber{7}  

\title{SIGNet: Intrinsic Image Decomposition by a Semantic and Invariant Gradient Driven Network for Indoor Scenes} 


\titlerunning{SIGNet}
%
\author{Partha Das\inst{1,3}\orcidlink{0000-0003-0112-3638} \and
Sezer Karaoğlu\inst{1,3} \and
Arjan Gijsenij\inst{2} \and
Theo Gevers\inst{1,3}}
\authorrunning{P. Das et al.}
%
\institute{CV Lab, University of Amsterdam, The Netherlands\and
AkzoNobel, The Netherlands \and
3DUniversum, Amsterdam, The Netherlands \\
\email{\{p.das, th.gevers\}@uva.nl} \& \email{s.karaoglu@3duniversum.com} \& \email{arjan.gijsenij@akzonobel.com} }

\maketitle

\begin{abstract}
Intrinsic image decomposition (IID) is an under-constrained problem. Therefore, traditional approaches use hand crafted priors to constrain the problem. However, these constraints are limited when coping with complex scenes. Deep learning-based approaches learn these constraints implicitly through the data, but they often suffer from dataset biases (due to not being able to include all possible imaging conditions).

In this paper, a combination of the two is proposed. Component specific priors like semantics and invariant features are exploited to obtain semantically and physically plausible reflectance transitions. These transitions are used to steer a progressive CNN with implicit homogeneity constraints to decompose reflectance and shading maps. 

An ablation study is conducted showing that the use of the proposed priors and progressive CNN increase the IID performance. State of the art performance on both our proposed dataset and the standard real-world IIW dataset shows the effectiveness of the proposed method. Code is made available~\href{https://github.com/Morpheus3000/SIGNet}{here.}

\end{abstract}

\section{Introduction}

An image can be defined as the combination of an object's colour and the incident light on it projected on a plane. Inverting the process of image formation is useful for many downstream computer vision tasks such as geometry estimation~\cite{henderson2019}, relighting~\cite{shu2017}, colour edits~\cite{Beigpour2011ObjectRB} and Augmented Reality (AR) insertion and interactions for applications like the Metaverse. The process of recovering the object colour (reflectance or albedo) and the incident light (shading) is known as Intrinsic Image Decomposition (IID). As the problem is ill-defined (with only one known), constraint-based approaches are explored to limit the solution space. For example, as an explicit gradient assumption, softer (or smoother) gradient transitions are attributed to shading transitions, while stronger (or abrupt) ones are related to reflectance transitions~\cite{Land1971}. Colour palette constraints in the form of sparsity priors and piece-wise consistency are also employed for reflectance estimation~\cite{Gehler2011}, \cite{Barron2015}. However, these approaches are based on strong assumptions of the imaging process and hence are limited in their applicability.

Implicit constraints, by means of deep learning-based methods, are proposed to expand previous approaches~\cite{Narihia2015}. For these methods, the losses implicitly formulate the constraints and are dependent on the training data. These methods learn a flexible representation based on training data which may lead to dataset biases.~\cite{Li2018ECCV} integrates multiple datasets to manage the dataset bias problem. However, introducing more datasets only acts as an expansion of the imaging distribution. Additionally, multiple purpose-built losses are needed to train the network. An alternative approach of combining constraints and deep learning is explored in~\cite{Fan2018} where edges are used as an additional constraint to guide the network. However, edges at image locations with strong illumination effects, like pronounced cast shadows, may lead to edge misclassification resulting in undesirable effects like shading-reflectance leakages. 

On the other hand,~\cite{Baslamisli2018ECCV} forgoes priors and specialised losses to leverage joint learning of related modalities. They explore semantic segmentation as a closely related task to IID, arguing that jointly learning the semantic maps provides the network information to jointly correct for reflectance-shading transitions. However, no explicit guidance or constraint between the semantics and reflectance are imposed. The network thus relies on learning the constraints from the ground truth semantic, reflectance and shading jointly. Moreover, only outdoor gardens are considered, where most natural classes (e.g., bushes, trees, and roses) contain similar colours (i.e., constrained colour distributions).

\begin{figure}
    \centering
    \includegraphics[width=0.9\linewidth]{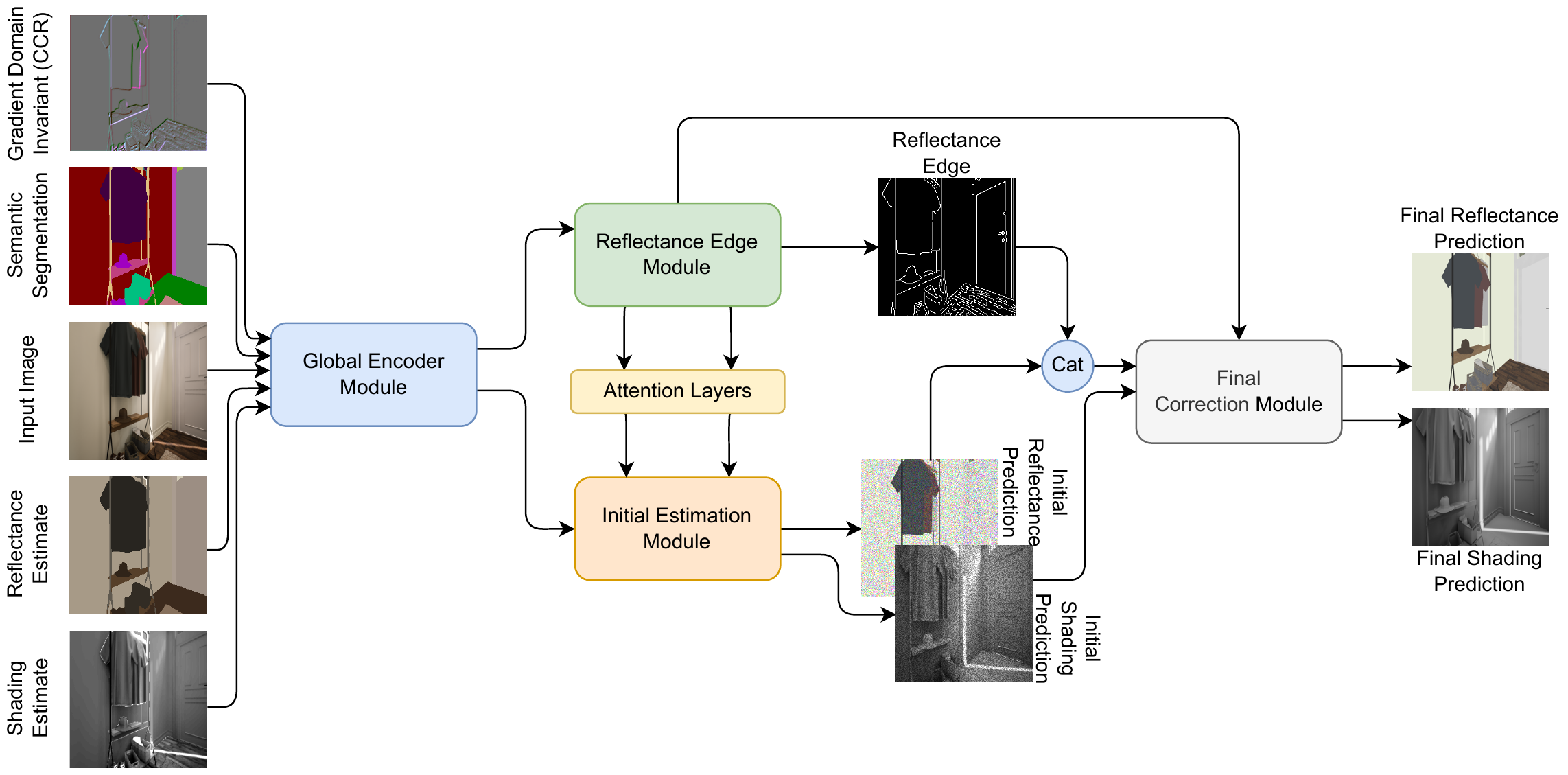}
    \caption{The proposed network overviews. The network consists of i) the global encoder module, ii) the reflectance edge module, iii) the initial estimation module, and iv) the final correction module. The final reflectance and shading outputs are used for all the evaluations. Please refer to the supplementary for more details. Images shown here are ground truth images, for illustrative purposes.}
    \label{fig:NetOverview}
\end{figure}

This paper exploits physical and statistical image properties for IID of indoor scenes. Illumination and geometry invariant descriptors~\cite{Gevers1999} yield physics-based cues to detect reflectance transitions, while statistical grouping of pixels in an image provides initial starting estimates for IID components. To this end, a combination of semantic and invariant transition constraints is proposed. Semantic transitions provide valuable information about reflectance transitions i.e., a change in semantics most likely matches a reflectance transition but not always the other way around (objects may consist of different colours). Illumination invariant gradients provide useful information about reflectance transitions but can be unstable (noisy) due to low intensity. Exploiting reflectance transition information on these two levels compensates each other and ensures a stronger guidance for IID. In addition, indoor structures, like walls and ceilings, are often homogeneously coloured. To this end, the semantic map can be used as an explicit homogeneous prior. This allows for integrating an explicit sparsity/piece-wise consistency (homogeneity) prior in the form of constant reflectance colour. 

In this paper, a progressive CNN is employed, consisting of two stages. The first stage of the network exploits the prior information to arrive at an initial estimation. This estimation is based on the semantics, the invariant guided boundaries, and sparsity constraints. The second stage of the network takes the initial estimation and fine-tunes it using the original image cues to disentangle the reflectance and shading maps while being semantically correct. This allows the network to separate the problem into two distinct solution spaces that build progressively on each other. In addition, it allows the network to learn a continuous representation that can extrapolate even when the priors contain errors. An overview of the proposed network is shown in the Fig.~\ref{fig:NetOverview}.

While deep learning networks have shown very good performance, they require high quality datasets. Traditional physical-based rendering methods are often time and resource intensive. Recently, these methods are more efficient i.e., real time on consumer hardware. Hence, a dataset of physical-based and photo-realistic rendered indoor images is provided. The synthetic dataset is used to train the proposed method.

In summary, our contributions are as follows:
\begin{itemize}
    \item \textbf{Algorithm}: An end-to-end semantic and physically invariant edge transition driven hybrid network is proposed for  intrinsic image decomposition of indoor scenes.
    \item \textbf{Insight}: The use of component specific priors outperforms learning from a single image.
    \item \textbf{Performance}: The proposed algorithm is able to achieve state-of-the-art performance on both synthetic and real-world datasets.
    \item \textbf{Dataset}: A new ray-traced and photo-realistic indoor dataset is provided.

\end{itemize}

\section{Related Works}

A considerable amount of effort has been put in exploring hand-crafted prior constraints for the problem of IID. ~\cite{Land1971} pioneered the field by assuming reflectance changes to be related to sharper gradient changes, while smoother gradients correspond to shading changes. Other priors have been explored like piece-wise constancy for the reflectance, and smoothness priors for shading~\cite{Barron2015}, textures~\cite{Gehler2011}. Constraints in the form of additional inputs have also been explored.~\cite{Lee2012} explores the use of depth as an additional input, while~\cite{Jeon2014} explores surface normals. Near infrared priors are used by~\cite{Cheng2019} to decompose non-local intrinsics. Humans in the loop is also studied by~\cite{Bonneel2014} and~\cite{Narihira2015-2}. However, these works mostly focus on single objects and do not generalise well to complete scenes.

In contrast to the use of explicit (hand-crafted) constraints, deep learning methods that implicitly learn (data-driven) specific constraints are also explored~\cite{Narihia2015}. ~\cite{Baslamisli2019} explores disentangling the shading component into direct and indirect shading.~\cite{Zhou2019} differentiates shading into illumination and surface normals in addition to reflectance.~\cite{Bell2014} uses a piece-wise constancy property of reflectances and employs Conditional Random Fields to perform IID.~\cite{Fan2018} shows that image edges contain information about reflectance edges and uses them as a guidance for the IID problem.~\cite{Li2018ECCV} reduces the solution space by using multiple task specific losses.~\cite{Sengupta2019} directly learns the inverse of the rendering function. Finally,~\cite{Baslamisli2018ECCV} forgoes losses and jointly learns semantic segmentation to implicitly learn a posterior on the IID, while~\cite{Saini2019} uses estimated semantic features as a support for an iterative competing formulation for IID. However, the above approaches do not explicitly integrate the physics-based image formation information and rely on the datasets containing a large set of imaging conditions. Hence, they may fall short for images containing extreme imaging conditions such as strong shadows or reflectance transitions. Large datasets~\cite{Li2018ECCV,roberts2021,Li2021CVPR} are proposed to train networks. Unfortunately, they are limited in their photo-realistic appearance.

Unlike IID, physics-based image formation priors have been explored in other tasks.~\cite{Finlayson1992} introduces Colour Ratios which are illumination invariant descriptors for objects.~\cite{Gevers1999} then introduces Cross Colour Ratios which are both geometric and illumination invariant reflectance descriptors.~\cite{Baslamisli2020} shows the applicability of the descriptors to the problem of IID. In contrast to previous methods, in this paper, a combination of explicit image formation-based priors and implicit intrinsic component property losses are explored.

\section{Methodology}

\subsection{Priors}

\paragraph{\textbf{Semantic Segmentation:}}~\cite{Baslamisli2018ECCV} shows that semantic segmentation provides useful information for the IID problem. However, components are jointly learned and hence their method lacks any explicit influence of the component's property. Since object boundaries correspond to reflectance changes such boundary information can serve as a useful global reflectance transition guidance for the network. Furthermore, homogeneous colour (i.e., reflectance) constraints (e.g., a wall has a uniform colour) can be imposed on the segmentation explicitly. To this end, in this paper, an off-the-self segmentation algorithm Mask2Former~\cite{cheng2021} is used to obtain segmentation maps.

\paragraph{\textbf{Invariant Gradient Domain:}} Solely using semantic regions as priors may cause the network to be biased to the regions generated by the segmentation method. To prevent such a bias, an invariant (edge) map is included as an additional prior to the network. In this work, Cross Colour Ratios (CCR)~\cite{Gevers1999} are employed. These are illumination invariants i.e., reflectance descriptors. Given an image $I$ with channels Red ($R$), Green ($G$) and Blue ($B$) and neighbouring pixels $p_1$ and $p_2$, CCR is defined by $M_{RG} = \;\frac{R_{p_1}\;G_{p_2}}{R_{p_2}\;G_{p_1}}\;$, $M_{RB} = \;\frac{R_{p_1}\;B_{p_2}}{R_{p_2}\;B_{p_1}}\;$ and $M_{GB} = \;\frac{G_{p_1}\;B_{p_2}}{G_{p_2}\;B_{p_1}}\;$ where, $R_{p_1}$, $G_{p_1}$ and $B_{p_1}$ are the red, green, and blue channel for pixel $p_1$. Descriptors $M_{RG}$, $M_{RB}$ and $M_{GB}$ are illumination free and therefore solely depending on reflectance transitions. Using the reflectance gradient as an additional prior allows the network to be steered by reflectance transitions. 

\paragraph{\textbf{Reflectance and Shading Estimates:}} Consider the simplified Lambertian~\cite{Shafer1985} image formation model: $I = R \times S$, where shading ($S$) is the scaling term on the reflectance component ($R$). Hence, for a given constant reflectance region, all the pixels are different shades of the same colour. In this way, the reflectance colour becomes a scale optimisation for which the pixel mean of a segment can be used: $\mathcal{M}_c = \sum^N_{n=1} {I^c_n}$ where, $\mathcal{M}_c$ is the channel-specific mean of the pixels. $\mathcal{M}_R$, $\mathcal{M}_G$ and $\mathcal{M}_B$ values are then spread within the region to obtain an initial starting point for reflectance colour based on the homogeneity constraint. Conversely, these values can be inverted using the image formation to obtain the corresponding scaled shading estimates. A CNN is then employed to implicitly learn the scaling for both priors. Additionally, since the mean of the segment does not consider textures, a deep learning method is proposed to compensate it by means of a dedicated correction module, see section~\ref{sec:net_arch}. The supplementary material provides more visuals for these priors.

\subsection{Network Architecture}\label{sec:net_arch}

The network consists of $4$ components: i) Global encoder blocks, ii) Reflectance edge Decoder, iii) Initial estimation decoder and iv) Final correction module. The network is trained end-to-end. The input to the network is an image and its corresponding segmentation obtained by Mask2Former~\cite{cheng2021}. The CCR, Reflectance and Shading estimates are computed from the input image for the respective encoder blocks. Additional details and visuals for the modules can be found in the supplementary materials.

\begin{figure}
    \centering
    \includegraphics[width=0.83\linewidth]{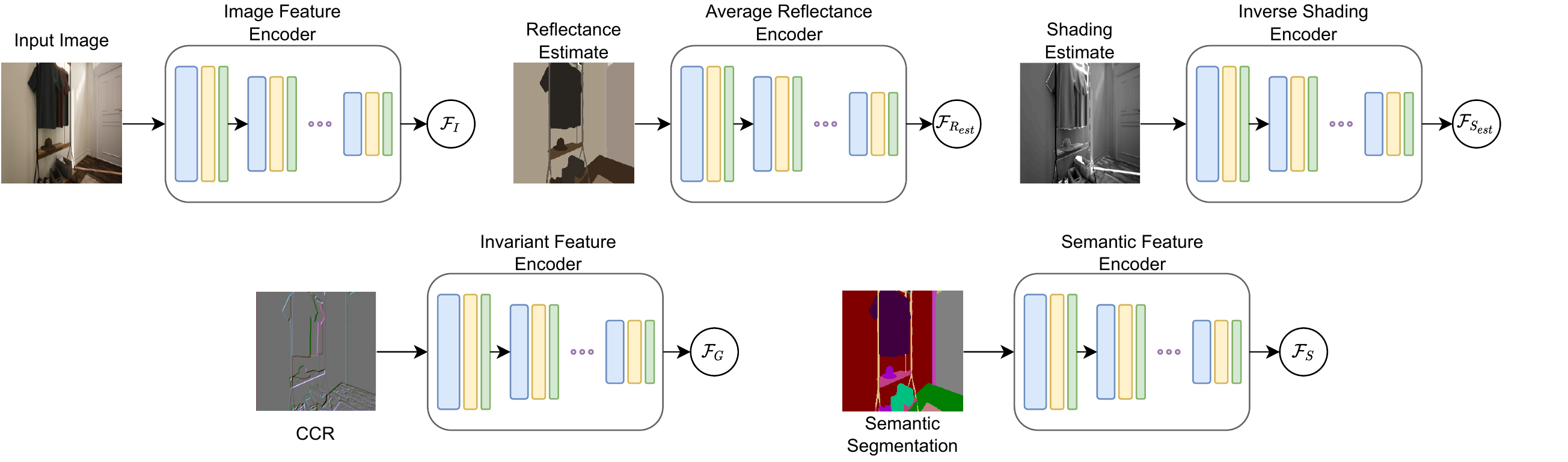}
    \caption{Overview of the global encoder module. Each of the inputs are provided with their independent encoders to enable modality specific feature learning. The respective features are used in the downstream decoders to provide component specific information for the network.}
    \label{fig:GlobalModule}
\end{figure}

\paragraph{\textbf{Global Encoder Module:}} The input image, the segmentation image, the average reflectance estimate, inverse shading estimate and the CCR images are encoded through their respective encoders. The encoders share the same configuration, but the intermediate features are independent of each other. The semantic features ($\mathcal{F}_{S}$) provide guidance for the general outlines of object boundaries, while the CCR features ($\mathcal{F}_{G}$) focus on local reflectance transitions, possibly including textures. Correspondingly, the average reflectance estimate features ($\mathcal{F}_{R_{est}}$) and the inverse shading estimate features ($\mathcal{F}_{S_{est}}$) provide a starting point for the reflectance and the shading estimation, respectively. Finally, the image features ($\mathcal{F}_{I}$) provide the network a common conditioning to learn the scaling and boundary transitions for the intrinsic components. Fig.~\ref{fig:GlobalModule} shows the overview of the module.

\begin{figure}
    \centering
    \includegraphics[width=0.83\linewidth]{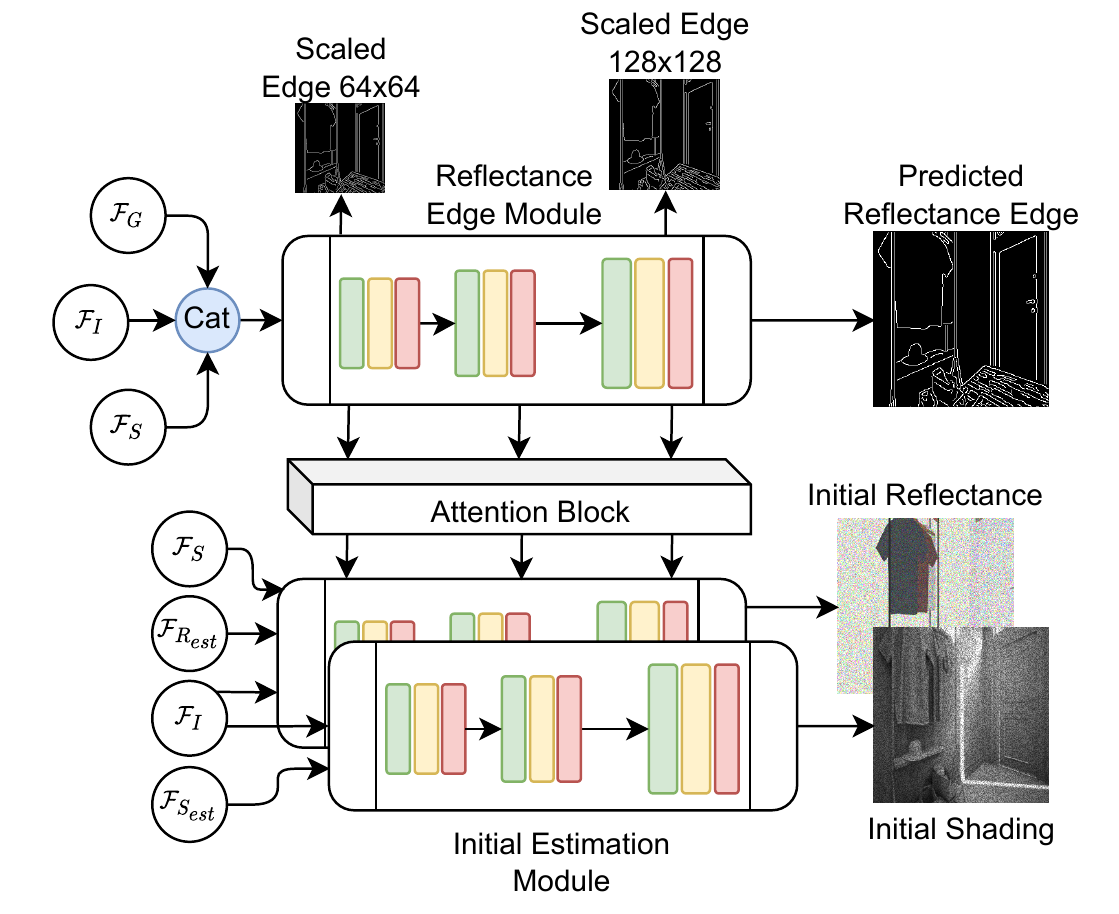}
    \caption{Overview of the reflectance edge and the attention guided initial estimation module. The edge module takes the image encoder, semantic encoder, and the invariant encoder feature to learn a semantically and physically guided reflectance transition. The edge features are then transferred through an attention block to the initial estimation decoder module. The reflectance decoder in this module takes the semantic encoder, image encoder and the average reflectance estimation features and input. The shading decoder correspondingly takes the image encoder along with the average shading estimation feature. Interconnections in the decoder allows the network to use reflectance cues for shading and vice versa.}
    \label{fig:EdgeDecoder}
\end{figure}

\paragraph{\textbf{Reflectance Edge Module:}} This sub-network decodes the reflectance edges of the given input. The decoded reflectance and edges are used as an attention mechanism to the initial estimation module to provide (global) region consistency. The features, $\mathcal{F}_{S}$ and $\mathcal{F}_{G}$ are concatenated with the image features $\mathcal{F}_{I}$ and passed on to the edge decoder. The semantic and CCR features provide object and reflectance transitions, respectively. The image features allow the network to disentangle reflectance from illumination edges. Corresponding skip connections from $\mathcal{F}_{I}$, $\mathcal{F}_{R_{est}}$ and $\mathcal{F}_{G}$ encoders are used to generate high frequency details. Scale space supervision, following~\cite{Xie2015}, is provided by a common deconvolution layer for the last $2$ layers, for scales of $64\times64$ and $128\times128$, yielding a scale consistent reflectance edge prediction. The ground truth edges are calculated by using a Canny Edge operation on the ground truth reflectance. Fig.~\ref{fig:EdgeDecoder} shows an overview of the module.

\paragraph{\textbf{Initial Estimation Module:}} The initial estimation decoder block focuses on learning the IID from the respective initial estimates of the intrinsic (Fig.~\ref{fig:EdgeDecoder}). It consists of two parallel decoders. The Reflectance decoder learns to predict the first estimation from  $\mathcal{F}_{I}$ and $\mathcal{F}_{R_{est}}$. The features are further augmented with the learned boundaries from the reflectance edge decoder passed through an attention layer~\cite{tang2020}. $\mathcal{F}_{S}$ is also passed to the decoder to guide global object transitions and acts as an additional attention. Similarly, the Shading decoder only receives $\mathcal{F}_{I}$ and $\mathcal{F}_{S_{est}}$, focusing on properties like smoother (shading) gradient changes. The reflectance and shading decoders are interconnected to provide an additional cue to learn an inverse of each other. Skip connections from the respective encoders to the decoders are also given. This allows the network to learn an implicit scaling on top of the average reflectance and the inverse shading estimation. The output at this stage is guided by transition and reflectance boundaries and may suffer from local inconsistencies like shading-reflectance leakages.

\begin{figure}
    \centering
    \includegraphics[width=\linewidth]{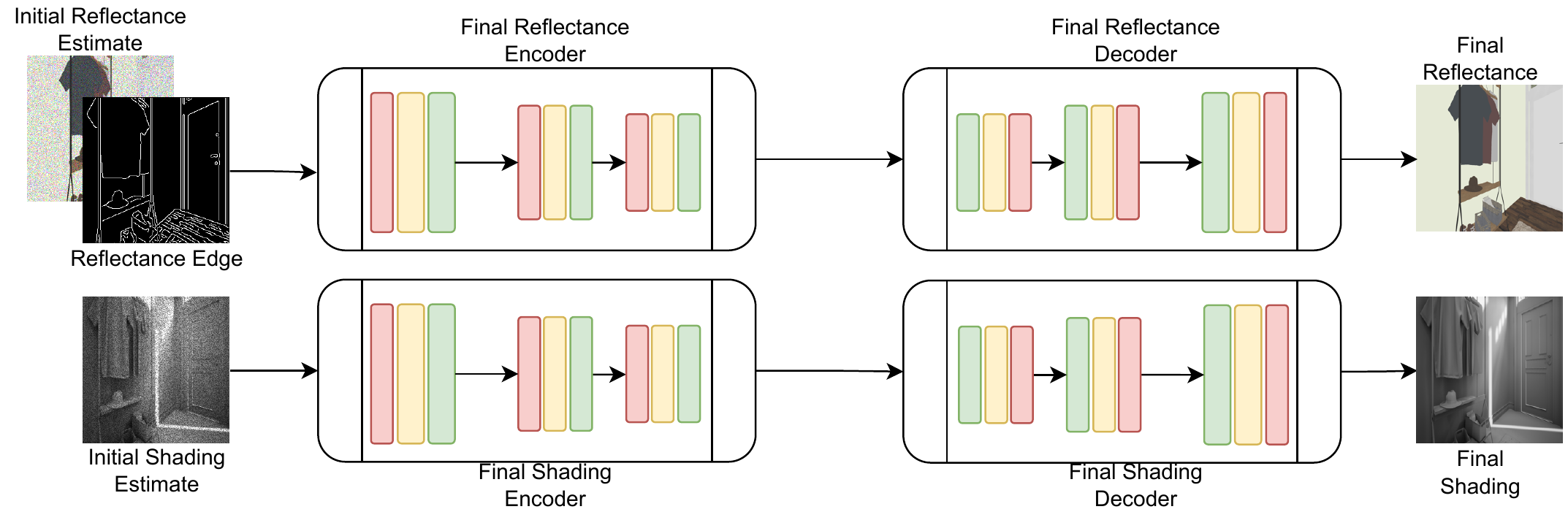}
    \caption{The final decoder module. The initial reflectance and shading estimates from the previous step are further corrected to obtain the final reflectance and shading. The encoder consists of an independent parallel reflectance and shading encoder. The reflectance encoder takes receives the initial reflectance and the reflectance edge as an input, while the shading encoder receives the initial shading. Two parallel decoders are used for reflectance and shading to obtain the final IID outputs.}
    \label{fig:FinalDecoder}
\end{figure}

\paragraph{\textbf{Final Correction Module:}} To deal with local inconsistencies, a final correction module is proposed. First, the reflectance edge from the edge decoder and the reflectance from the previous decoder is concatenated and passed through a feature calibration layer. This allows the network to focus on local inconsistencies guided by global boundaries. The output is then passed through a final reflectance encoder. The shading from the previous module is similarly passed through another encoder block. The output of these two encoders is then passed through another set of parallel decoders for the final reflectance and shading output. Since the reflectance and shading from the previous block is already globally consistent, this decoder acts as a localised correction. To constrain the corrections to local homogeneous regions, skip connections (through attention layers) of encoded reflectance edge features are provided to the decoders. In this way, the network limits the corrections to the local homogeneous regions and recover local structures like textures. Skip connections from the respective reflectance and shading encoders are provided to include high frequency information transfer. The reflectance and shading features in the decoder are shared within each other to enforce an implicit image formation model. Fig.~\ref{fig:FinalDecoder} shows the overview of the module.

\subsection{Dataset}

Unreal Engine~\cite{unrealengine} is used to generate a dataset suited for the task. The rendering engine supports physically based rendering, with real-time raytracing (RTX) support. The engine first calculates the intrinsic components from the various material and geometry property of the objects making up the scene. Then, the illumination is physically simulated through ray tracing and lighting is calculated. Finally, all these results are combined to render the final image. Since the engine calculates the intrinsic components,  ground truth intrinsic is recovered using the respective buffer. The dataset consists of dense reflectance and shading ground-truths. The network learns the inversion of this process.

\begin{figure}
    \centering
    \includegraphics[width=\linewidth]{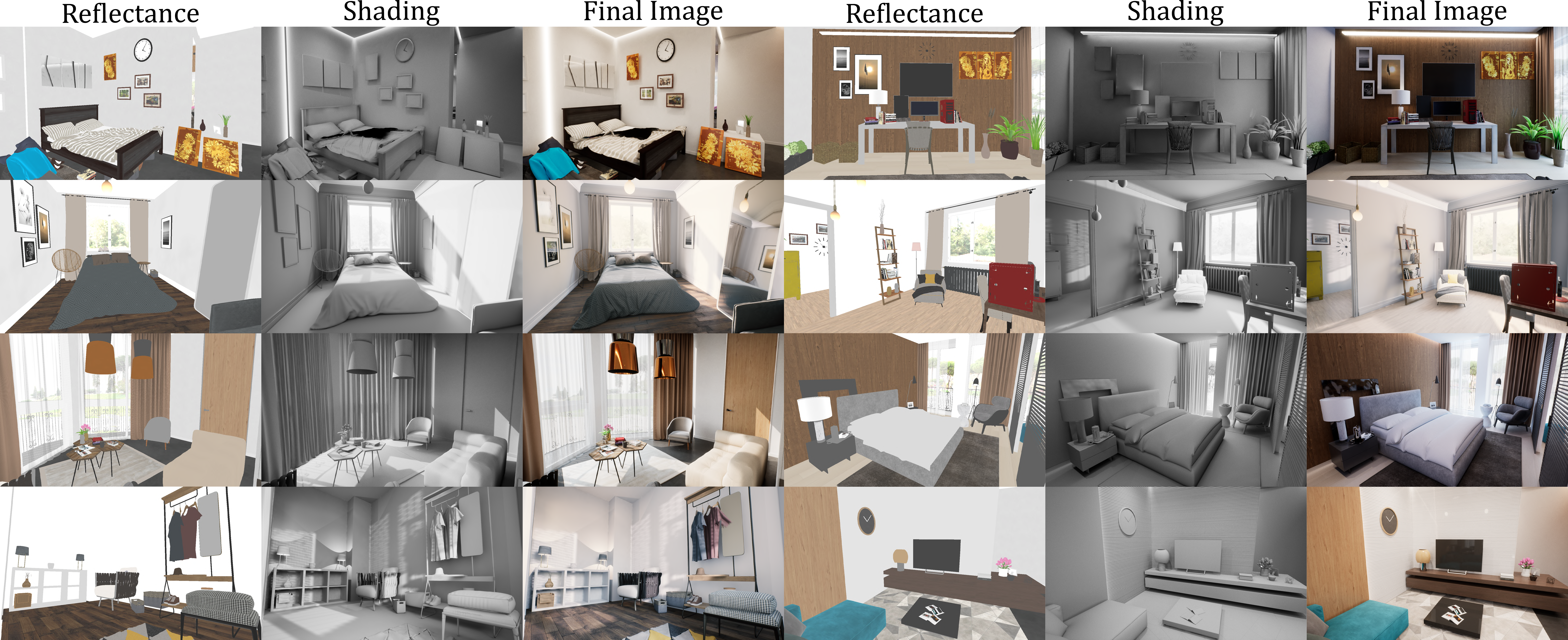}
    \caption{Samples from the proposed dataset. The dataset comes with the corresponding dense reflectance and shading maps. The dataset consists of various everyday objects and lighting, containing both near local light sources, like lamps, and more global light sources like sunlight and windows.}
    \label{fig:dataset}
\end{figure}

Assets from the unreal marketplace are used to generate the dataset. These assets are professionally created to be photo realistic. $5000$ images are generated of which $4000$ images are used for training, and $1000$ are used for validation and testing. To evaluate the generalisability of the network, Intrinsic Images in the Wild (IIW)~\cite{Bell2014} is used as a real-world test. Fig.~\ref{fig:dataset} shows a number of samples from the dataset. The dataset generated is comparatively small. However, the purpose of the dataset is that the network learns an efficient physics guided representation, rather than a dataset dependent one. The pretrained model and the dataset will be made available.

\subsection{Loss Functions and Training details}\label{sec:loss_training}

MSE loss is applied for each output of the network: (i) Initial estimation loss ($\mathcal{L}_e$ \& $\mathcal{L}_i$) and (ii) Final correction loss ($\mathcal{L}_f$). $\mathcal{L}_e$ is the loss applied on the scale space reflectance edge. $\mathcal{L}_i$ is the loss on the reflectance and shading output from the initial estimation module. Additional losses are also applied on the reflectance and shading output from the final correction module. This reflectance and shading are also combined and compared with the input image for a reconstruction loss. These $3$ losses are collected in the term $\mathcal{L}_f$. An invariance loss $\mathcal{L}_{Norm}$ is added between the normalised $RGB$ and the prediction of the network for each segment. A Total Variation (TV) loss ($\mathcal{L}_{TV}$) is included to deal with the assumption that large indoor classes like walls and ceilings are homogeneously coloured. This loss is only applied to ceilings and wall pixels and minimises the TV between the prediction and the ground truth reflectance. Finally, to encourage perceptually consistent and sharper textures, a perceptual and dssim loss are included and grouped as $\mathcal{L}_{\delta}$. The final loss term to minimise for the network thus becomes:

\begin{equation}
\begin{aligned}
    \label{eq:total}
    \mathcal{L} = \lambda_e \; \mathcal{L}_e + \lambda_i \; \mathcal{L}_i + \mathcal{L}_f \\
    + \mathcal{L}_{Norm} + \mathcal{L}_{TV} + \mathcal{L}_{\delta}
\end{aligned}
\end{equation}

\noindent where $\lambda_e$ and $\lambda_i$ are weighting terms for the edge and initial estimation losses. They are empirically set to $0.4$ and $0.5$, respectively. The network is trained for 60 epochs, with a learning rate of $2e-4$ and the Adam~\cite{kingma2014} optimiser. Please refer to the supplementary materials for more details.

\section{Experiments}

\subsection{Ablation Study}

To study the influence of different architecture components and losses, an ablation study is conducted. For a fair evaluation, the ablation study is performed on the test-set of the rendered dataset. For all the ablations, all hyper-parameters are kept constant. The results of the ablation study are presented in table~\ref{tab:ablation}.

\begin{table}
\centering
\caption{Ablation study for the proposed network. For each experiment, the respective parts of the network are modified. All the experiments are conducted on the same test and train split of the proposed dataset. All the applicable hyper-parameters are kept constant.}
\resizebox{0.8\textwidth}{!}{%
\begin{tabular}{c|ccc|ccc|}
\cline{2-7}
 & \multicolumn{3}{c|}{Reflectance} & \multicolumn{3}{c|}{Shading} \\ \cline{2-7} 
 & \multicolumn{1}{c|}{MSE} & \multicolumn{1}{c|}{LMSE} & DSSIM & \multicolumn{1}{c|}{MSE} & \multicolumn{1}{c|}{LMSE} & DSSIM \\ \hline
\multicolumn{1}{|c|}{w/o Final Correction} & \multicolumn{1}{c|}{0.0029} & \multicolumn{1}{c|}{0.0020} & 0.0225 & \multicolumn{1}{c|}{0.0044} & \multicolumn{1}{c|}{0.0035} & 0.0276 \\ \hline
\multicolumn{1}{|c|}{w/o Priors} & \multicolumn{1}{c|}{0.0105} & \multicolumn{1}{c|}{0.0047} & 0.0444 & \multicolumn{1}{c|}{0.0054} & \multicolumn{1}{c|}{0.0034} & 0.0399 \\ \hline
\multicolumn{1}{|c|}{w Canny Edges} & \multicolumn{1}{c|}{0.0032} & \multicolumn{1}{c|}{0.0037} & 0.0229 & \multicolumn{1}{c|}{0.0031} & \multicolumn{1}{c|}{0.0049} & 0.0293 \\ \hline
\multicolumn{1}{|c|}{w/o Average Estimates} & \multicolumn{1}{c|}{0.0030} & \multicolumn{1}{c|}{0.0023} & 0.0232 & \multicolumn{1}{c|}{0.0041} & \multicolumn{1}{c|}{0.0043} & 0.0267 \\ \hline
\multicolumn{1}{|c|}{w/o Reflectance Edge Module} & \multicolumn{1}{c|}{0.0097} & \multicolumn{1}{c|}{0.0156} & 0.3254 & \multicolumn{1}{c|}{0.0033} & \multicolumn{1}{c|}{0.0061} & 0.0270 \\ \hline \hline
\multicolumn{1}{|c|}{No DSSIM Loss} & \multicolumn{1}{c|}{0.0131} & \multicolumn{1}{c|}{0.0240} & 0.3704 & \multicolumn{1}{c|}{0.0041} & \multicolumn{1}{c|}{0.0055} & 0.1488 \\ \hline 
\multicolumn{1}{|c|}{No Perceptual Loss} & \multicolumn{1}{c|}{0.0032} & \multicolumn{1}{c|}{0.0022} & 0.0289 & \multicolumn{1}{c|}{0.0032} & \multicolumn{1}{c|}{0.0038} & 0.0285 \\ \hline
\multicolumn{1}{|c|}{No Invariant \& Homogeneity Loss} & \multicolumn{1}{c|}{0.0032} & \multicolumn{1}{c|}{0.0027} & 0.0288 & \multicolumn{1}{c|}{\textbf{0.0024}} & \multicolumn{1}{c|}{\textbf{0.0024}} & 0.0318 \\ \hline \hline
\multicolumn{1}{|c|}{Proposed} & \multicolumn{1}{c|}{\textbf{0.0026}} & \multicolumn{1}{c|}{\textbf{0.0018}} & \textbf{0.0219} & \multicolumn{1}{c|}{0.0030} & \multicolumn{1}{c|}{0.0033} & \textbf{0.0252} \\ \hline
\end{tabular}%
}
\label{tab:ablation}
\end{table}

\paragraph{\textbf{Influence of final correction module:}} In this experiment, the influence of the final correction module is studied. The output from the initial estimation decoder is taken as the final output. 

From the results, it is shown that the final correction module helps in improving the outputs. The improvement in the DSSIM metric for both  components shows that the final correction module is able to deal with structural artefacts.

\paragraph{\textbf{Influence of priors:}} The influence of all the priors is studied in this ablation. The additional priors are removed, and the network only receives the image as an input. All network structures are kept the same. This setup studies if the network can disentangle the additional information from the input image without any specific priors.

Removing all priors makes the network to perform worse for all metrics. In this setting, the network only uses the image to derive both the reflectance and shading changes. This is challenging for strong illumination effects. This shows that the priors are an important source of information enabling a better disentanglement between intrinsic components.

\paragraph{\textbf{Influence of specialised edges:}} This experiment studies the need of specialised edges obtained from the semantic transition boundaries and invariant features. The edges obtained from the input image are provided to the network. The study focuses on whether the network can distinguish between reflectance, geometry, and shadow edges directly from the image.

From the results, it is shown that using image edges is not sufficient. Image edges can be ambiguous due to the presence of shadow edges. However, the performance is still better than using the image as the only input, showing that edges yield, to a certain extent, useful transition information.

\paragraph{\textbf{Influence of reflectance and shading estimate priors:}} In this experiment, the efficacy of the statistic-based homogeneous reflectance and the inverted shading estimate is studied. 

Removing the average reflectance and shading estimates degrades the performance. With the priors of the estimates, the network can use its learning capacity to deal with the scaling of the initial estimation to obtain the correct IID. The network needs to learn the colour as well as the scaling within the same learning capacity.

\paragraph{\textbf{Influence of reflectance edge guidance module:}} For this experiment, the edge guidance module is removed. As such, the network is then forced to learn the attention and the reflectance transition boundaries implicitly as part of the solution space. 

Removing the reflectance edge module results in the second worse result. This shows that, apart from the priors, the ability to use those features to learn a reflectance transition, is useful. It is shown that without such a transition guidance, the network is susceptible of misclassifying shadow edges as reflectance transitions. Furthermore, it is shown that without this module, the reflectance performance suffers more than the shading performance. Hence, using a learned edge guidance allows the network to be more flexible and better able to distinguish between true reflectance transitions.

\paragraph{\textbf{Influence of different losses:}} The influence of the different losses is studied in this experiment. For each sub-experiment, the same proposed structure is used, and the respective losses are selectively turned off.

From the results, it is shown that the DSSIM loss contributes to a large extend, to the performance, because this loss penalises perceptual variations like contrast, luminance, and structure. As such, by removing the supervision, the network learns an absolute difference which is not expressive to smaller spatial changes. Similar trend of performance decrease is shown when removing the perceptual and homogeneity losses. This is expected since both losses contribute to region consistency. With the addition of the losses on the reflectance, the shading values suffer slightly. However, structurally they perform better when including the losses, as shown by the DSSIM metric. This indicates that applying such a loss helps not only to achieve a better reflectance value, but it also jointly improves shading, resulting in sharper outputs.

\subsection{Comparison to State of the Art}

\begin{figure}
    \centering
    \includegraphics[width=\linewidth]{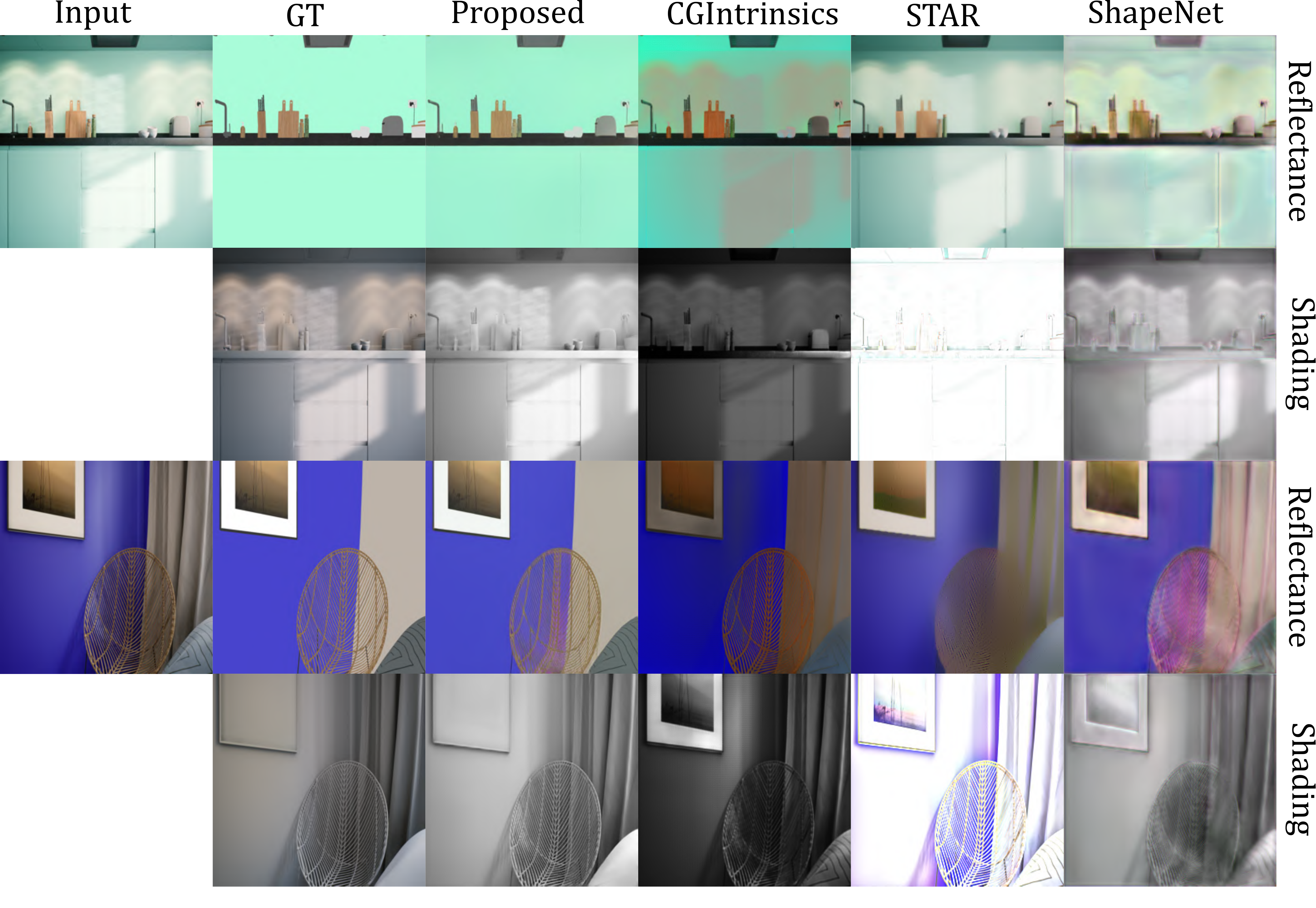}
    \caption{Comparison of the proposed to baseline methods. It is shown that the proposed method is able to better disentangle the illumination effect. In comparison, CGIntrinsics, which has comparable results on the WHDR SoTA, suffers from discolouration. STAR misses the illumination while ShapeNet suffers from artefacts.}
    \label{fig:baseline}
\end{figure}

\paragraph{\textbf{On the proposed Dataset}:} To study the influence of the dataset, the proposed network is compared to baseline algorithm's performance. For these experiments, the standard, MSE, LMSE and the DSSIM metric are used. The baselines are chosen based on their performance of the Weighted Human Disagreement Rate (WHDR), widely used in the literature. Hence,~\cite{Li2018ECCV} is chosen as a baseline.~\cite{Zhou2019} does not provide any publicly available code, hence is not included. Although~\cite{Fan2018} is the state of the art, their provided code generates errors when trying to run on custom datasets and hence is not used for comparison. For completeness,~\cite{Xu2020} and ~\cite{Shi2017} is also compared.~\cite{Xu2020} uses an optimization-based method based on the pioneering Retinex model. Since it is a purely physical constraint-based model, it is included for comparison. For a fair comparison, methods focusing on indoors are used.~\cite{Baslamisli2018ECCV} assumes outdoor settings and requires semantic ground truths to train and hence is not included. For all the networks, they are retrained on the dataset that is proposed in this paper, using the optimum hyperparamters as mentioned in the respective publication. The results are shown in table~\ref{tab:baseline} and figure~\ref{fig:baseline}

\begin{table}
\centering
\caption{Comparison to the baseline methods on the proposed dataset. It is shown that the proposed method outperforms all other methods.}
\resizebox{0.7\textwidth}{!}{%
\begin{tabular}{c|ccc|ccc|}
\cline{2-7}
 & \multicolumn{3}{c|}{Reflectance} & \multicolumn{3}{c|}{Shading} \\ \cline{2-7} 
 & \multicolumn{1}{c|}{MSE} & \multicolumn{1}{c|}{LMSE} & DSSIM & \multicolumn{1}{c|}{MSE} & \multicolumn{1}{c|}{LMSE} & DSSIM \\ \hline
\multicolumn{1}{|c|}{ShapeNet~\cite{Shi2017}} & \multicolumn{1}{c|}{0.0084} & \multicolumn{1}{c|}{0.0133} & 0.1052 & \multicolumn{1}{c|}{0.0065} & \multicolumn{1}{c|}{0.0129} & 0.1862 \\ \hline
\multicolumn{1}{|c|}{STAR~\cite{Xu2020}} & \multicolumn{1}{c|}{0.0304} & \multicolumn{1}{c|}{0.0166} & 0.1180 & \multicolumn{1}{c|}{0.0290} & \multicolumn{1}{c|}{0.0128} & 0.1572 \\ \hline
\multicolumn{1}{|c|}{CGIntrinsics~\cite{Li2018ECCV}} & \multicolumn{1}{c|}{0.0211} & \multicolumn{1}{c|}{0.0156} & 0.0976 & \multicolumn{1}{c|}{0.0848} & \multicolumn{1}{c|}{0.0577} & 0.2180 \\ \hline
\multicolumn{1}{|c|}{Proposed} & \multicolumn{1}{c|}{\textbf{0.0026}} & \multicolumn{1}{c|}{\textbf{0.0018}} & \textbf{0.0219} & \multicolumn{1}{c|}{\textbf{0.0030}} & \multicolumn{1}{c|}{\textbf{0.0033}} & \textbf{0.0252} \\ \hline
\end{tabular}%
}
\label{tab:baseline}
\end{table}

From the table it is shown that our proposed model is able to provide the highest scores. From the figure, the baselines suffer from strong  illumination effects. CGIntrinsics discolours the regions while STAR mostly fails. ShapeNet, suffers from artefacts and colour variations around the illumination regions. In comparison, the proposed network is able to recover from such effects.

\begin{figure}
    \centering
    \includegraphics[width=\linewidth]{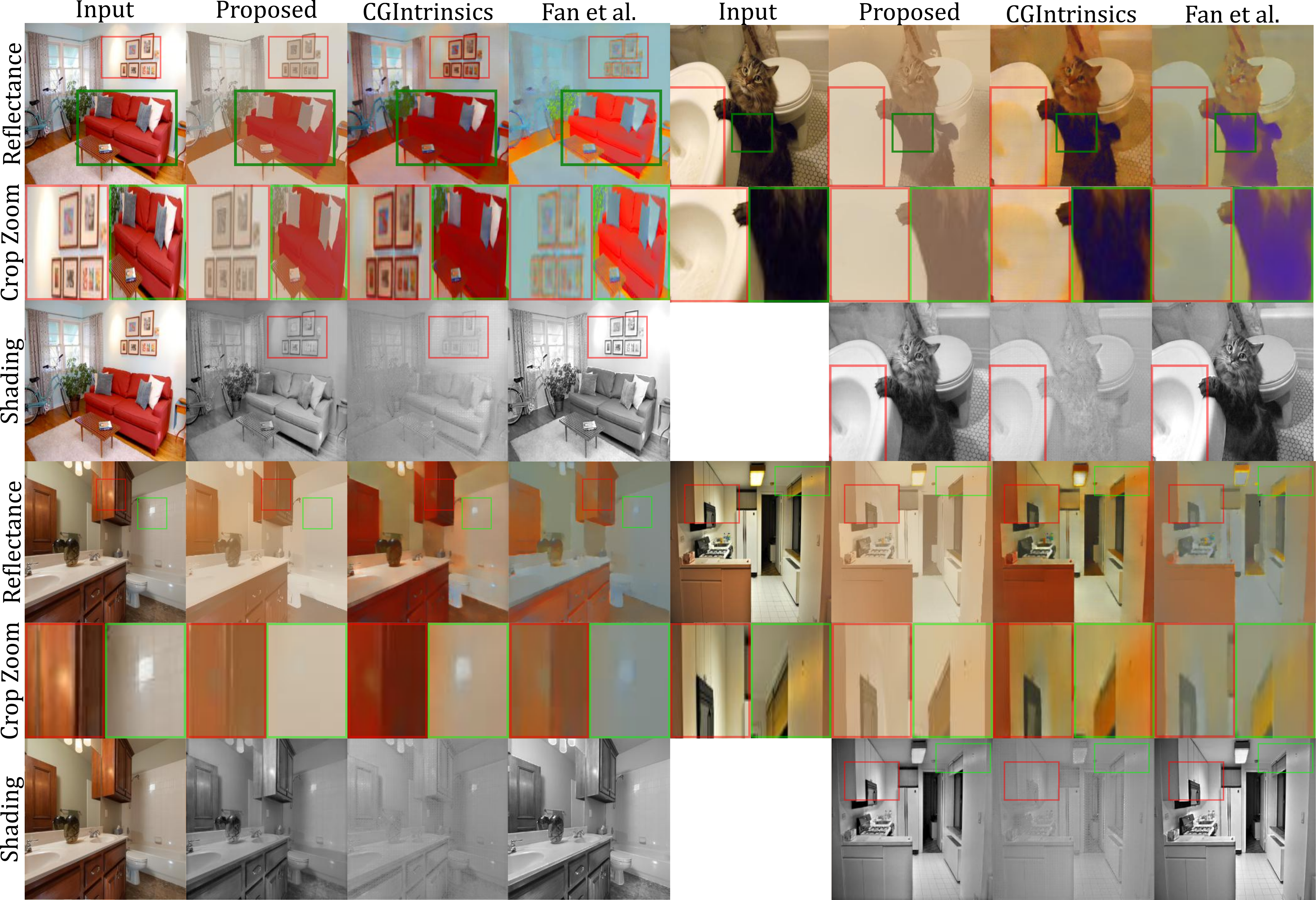}
    \caption{Visual results on the IIW test set. Compared to CGIntrinsics~\cite{Li2018ECCV} and Fan ~\etal~\cite{Fan2018}, the proposed method disentangles better the shading and highlights (highlighted in red boxes), showing a smoother reflectance. Both CGIntrinsics and~\cite{Fan2018} are unable to remove the highlights from the reflectance, resulting in discolouration. They are also susceptible to reflectance colour change as be seen on the cat and furniture (highlighted green boxes). The proposed method is able to better retain the original colour in the reflectance.}
    \label{fig:iiw_sota}
\end{figure}

\paragraph{\textbf{On IIW~\cite{Bell2014}}:} The proposed network is finetuned on the IIW dataset and compared to the baselines. The training and testing splits are used as specified in the original publication. For the baselines, the numbers are obtained from the respective original publications. The results are shown in table~\ref{tab:iiw_sota} and visuals in figure~\ref{fig:iiw_sota}.

\begin{table}
\centering
\caption{Baseline comparison for the IIW dataset. Results marked with * are postprocessed with a guided filter~\cite{Nestmeyer2016}}
\resizebox{0.4\textwidth}{!}{%
\begin{tabular}{|c|c|}
\hline
Methods & WHDR (mean) \\ \hline
Direct Intrinsics~\cite{Narihia2015} & 37.3 \\ \hline
Color Retinex~\cite{Grosse2009} & 26.9 \\ \hline
Garces~\etal~\cite{Garces2012} & 25.5 \\ \hline
Zhao~\etal~\cite{Zhao2012} & 23.2 \\ \hline
IIW~\cite{Bell2014} & 20.6 \\ \hline
Nestmeyer~\etal~\cite{Nestmeyer2016} & 19.5 \\ \hline
Bi~\etal~\cite{Bi2015} & 17.7 \\ \hline
Sengupta~\etal~\cite{Sengupta2019} & 16.7 \\ \hline
Li~\etal~\cite{Li2020} & 15.9 \\ \hline
CGIntrinsics~\cite{Li2018ECCV} & 15.5 \\ \hline
GLoSH~\cite{Zhou2019} & 15.2 \\ \hline
Fan~\etal~\cite{Fan2018} & 15.4 \\ \hline
Proposed & 15.2 \\ \hline \hline
CGIntrinsics~\cite{Li2018ECCV}* & 14.8 \\ \hline
GLoSH~\cite{Zhou2019}* & 14.6 \\ \hline
Fan~\etal~\cite{Fan2018}* & 14.5 \\ \hline
Proposed* & \textbf{13.9} \\ \hline
\end{tabular}%
}
\label{tab:iiw_sota}
\end{table}

The IIW dataset does not contain dense ground truth and hence is only finetuned with the ordinal loss. A guided filter~\cite{Nestmeyer2016} is used to further improve the results. Overall, our proposed method is on par with GLoSH~\cite{Zhou2019} which is the best performing method without any post filtering. However, they need both lighting and normal information as supervision, while the proposed method is trained with just reflectance and shading, along with a smaller dataset ($58,949$ images of~\cite{Zhou2019} vs. $5000$ of the proposed method). For the filtered results, the proposed method is able to achieve a comfortable lead compared to the current best of 14.5 obtained by~\cite{Fan2018}, showing the efficiency of the current model.

\section{Conclusions}

In this paper, an end-to-end prior driven approach for indoor scenes has been proposed for the task of intrinsic image decomposition. Reflectance transitions and invariant illuminant descriptors has been used to guide the reflectance decomposition. Image statistics-based priors have been used to provide the network a starting point for learning. To integrate explicit homogeneous constraints, a progressive CNN was used. To train the network, a custom physically rendered dataset was proposed. 

An extensive ablation was performed to validate the proposed network showing that: i) using explicit reflectance transition priors helps the network to achieve an improved intrinsic image decomposition, ii) image statistics-based priors are helpful for simplifying the problem and, iii) the proposed method attains sota performance for the standardised real-world dataset IIW.

\clearpage



\section{Supplementary: Network Architecture Details}

\begin{center}
    \begin{figure}
        \centering
        \hspace{2cm}
        \includegraphics[width=0.55\linewidth]{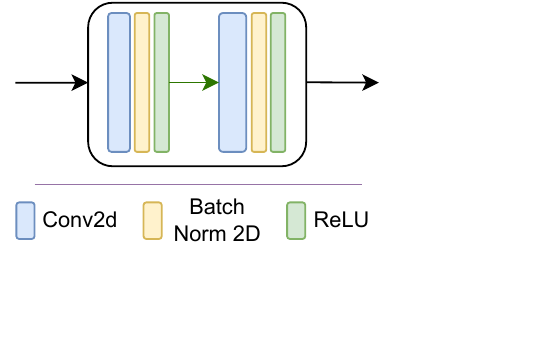}
        \vspace{-1cm}
        \caption{Encoder structure. Each encoder structure consists of $2$ groups of convolutions, followed by a batch normalization and a ReLU non-linearity layer.}
        \label{fig:encoder_structure}
    \end{figure}
\end{center}

\begin{center}
    \begin{figure}
        \centering
        \hspace{2cm}
        \includegraphics[width=0.55\linewidth]{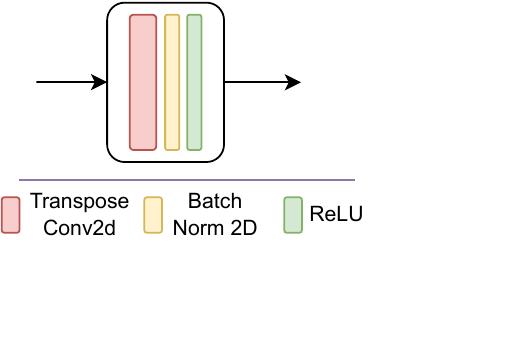}
        \vspace{-1cm}
        \caption{Decoder structure. Each decoder structure consists of a transposed convolution followed by a batch normalization and a ReLU non-linearity.}
        \label{fig:decoder_structure}
    \end{figure}
\end{center}

\begin{figure}
    \centering
    \includegraphics[width=0.6\linewidth]{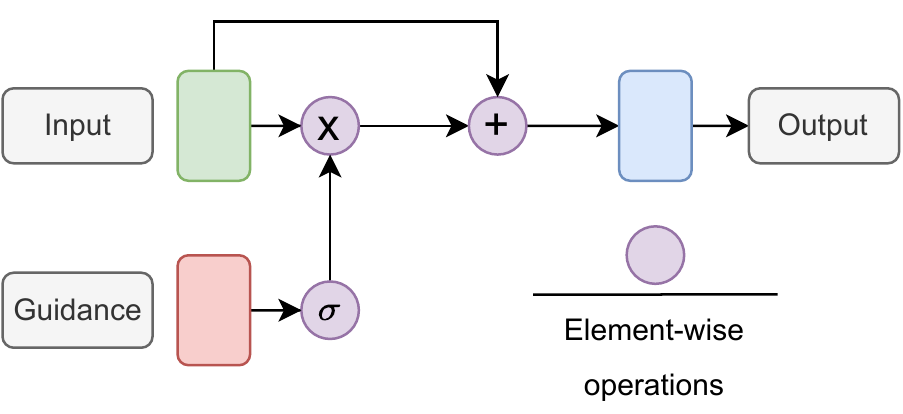}
    \caption{Attention layer. The layer receives two inputs, the guidance map, and the input over which the attention is applied. All the operations are elementwise, and the output is the same spatial and channel dimensions as the input.}
    \label{fig:attention_structure}
\end{figure}

The network consists of $3$ basic modules: i) encoder, ii) decoder and iii) attention. Figs~\ref{fig:encoder_structure}, \ref{fig:decoder_structure} and~\ref{fig:attention_structure} visualise these modules. 

These structures are used iteratively to construct the larger modules. 
\paragraph{\textbf{Global Encoder Module}:} All encoders share the same structure. Table~\ref{tab:encoder_def} lists the configuration.

\begin{table}
\centering
\caption{Overview of our encoder configuration used for each encoder module.}
\resizebox{0.8\textwidth}{!}{%
\begin{tabular}{|l|l|l|l|}
\hline
Name       & \textbf{Layer} & \textbf{Kernel Size, Stride, Padding} & \textbf{Output Size} \\ \hline
Input      & conv1 & 3x3x64, 1, 1                 & 256x256x64  \\ \hline
           & conv1 & 3x3x64, 1, 1                 & 256x256x64  \\ \hline
           & conv2 & 3x3x64, 2, 1                 & 128x128x64  \\ \hline
           & conv2 & 3x3x128, 1, 1                & 128x128x128 \\ \hline
           & conv3 & 3x3x128, 2, 1                & 64x64x128   \\ \hline
           & conv3 & 3x3x256, 1, 1                & 64x64x256   \\ \hline
           & conv4 & 3x3x256, 2, 1                & 32x32x256   \\ \hline
           & conv4 & 3x3x512, 1, 1                & 32x32x512   \\ \hline
           & conv5 & 3x3x512, 2, 1                & 16x16x512   \\ \hline
Bottleneck & conv5 & 3x3x512, 1, 1                & 16x16x512   \\ \hline
\end{tabular}%
}
\label{tab:encoder_def}
\end{table}

\paragraph{\textbf{Reflectance Edge Module}:} The configuration for the reflectance edge decoder module is listed in table~\ref{tab:reflec_edge_def}. The features being multiplied and added denote skip connections from the respective encoders. All features are first depth-wise concatenated before being passed through the decoder structures. The module has $3$ scaled outputs: $256\times256$, $128\times128$ and $64\times64$.

\begin{table}
\centering
\caption{Overview of the configuration for the edge decoders. The summations and product represent skip connections.}
\resizebox{0.8\textwidth}{!}{%
\begin{tabular}{|l|l|l|l|}
\hline
\textbf{Name} & \textbf{Layer} & \textbf{Kernel Size, Stride, Padding} & \textbf{Output Size} \\ \hline
BottleNeck & deconv1 & 4x4x(512 * 3), 2, 1             & 32x32x512   \\ \hline
           & deconv2 & 4x4x(512 * 4), 2, 1 & 64x64x512   \\ \hline
           & deconv3 & 4x4x(512 + (256 * 3)), 2, 1 & 128x128x256 \\ \hline
           & deconv4 & 4x4x(256 + (128 * 3)), 2, 1 & 256x256x128 \\ \hline
           & conv   & 3x3x(128 + (64 * 3)), 1, 1   & 256x256x64  \\ \hline
Full Edge 256     & conv   & 3x3x3, 1, 1                     & 256x256x3   \\ \hline
Edge Output 64     & conv   & 3x3x3, 1, 1                     & 256x256x3   \\ \hline
Edge Output 128     & conv   & 3x3x3, 1, 1                     & 256x256x3   \\ \hline
\end{tabular}%
}
\label{tab:reflec_edge_def}
\end{table}

\paragraph{\textbf{Initial Estimation Module}:} Tables~\ref{tab:init_reflec_def} and~\ref{tab:init_shading_def} show the module configuration. The decoder is a joint decoder in the style of~\cite{Shi2017}, i.e., the reflectance decoder blocks use the previous shading decoder outputs together with the previous reflectance decoder block outputs (denoted by $*$). The reflectance decoder receives the encoder features from the image encoder, the reflectance estimate encoder, and the semantic encoder as skip connections (denoted by $+$). Additionally, the edge decoder features, and the current reflectance decoder outputs are passed through the attention layer before being passed to the next block. Similarly, the shading decoder blocks receives shading estimate encode and the image encoder features as skip connections.

\begin{table}
\centering
\caption{Overview of the configuration for the initial reflectance estimation module decoder}
\resizebox{0.8\textwidth}{!}{%
\begin{tabular}{|l|l|l|l|}
\hline
\textbf{Name} & \textbf{Layer} & \textbf{Kernel Size, Stride, Padding}                                                            & \textbf{Output Size} \\ \hline
BottleNeck & deconv1   & 4x4x(512 * 3), 2, 1       & 32x32x512   \\ \hline
              & Attention      & reflect edge (re) \& deconv1 & 32x32x512            \\ \hline
           & deconv2   & 4x4x(512 * 2 + (512 + 512 + 512)), 2, 1 & 64x64x512   \\ \hline
           & Attention & re deconv2 \& deconv2 & 64x64x512   \\ \hline
           & deconv3   & 4x4x(512 * 2 + (256 + 256 + 256)), 2, 1 & 128x128x256 \\ \hline
           & Attention & re deconv3 \& deconv3  & 128x128x256 \\ \hline
           & deconv4   & 4x4x(256 * 2 + (128 + 128 + 128)), 2, 1 & 256x256x128 \\ \hline
           & Attention & re deconv4 \& decovn4  & 256x256x128 \\ \hline
           & conv6     & 3x3x(128 * 2 + (64 + 64 + 64)), 1, 1  & 256x256x64  \\ \hline
           & conv6     & 3x3x3, 1, 1               & 256x256x3   \\ \hline
Initial Reflectance Output     & Attention & re output \& conv6    & 256x256x3   \\ \hline
\end{tabular}%
}
\label{tab:init_reflec_def}
\end{table}

\begin{table}
\centering
\caption{Overview of the configuration for the initial shading estimation module decoder. The attention layer outputs are reused from the reflectance decoder.}
\resizebox{0.8\textwidth}{!}{%
\begin{tabular}{|l|l|l|l|}
\hline
\textbf{Name} & \textbf{Layer} & \textbf{Kernel Size, Stride, Padding}                                                            & \textbf{Output Size} \\ \hline
BottleNeck & deconv1   & 4x4x(512 * 2), 2, 1       & 32x32x512   \\ \hline
              & Attention      & reflect edge (re) \& deconv1 & 32x32x512            \\ \hline
           & deconv2   & 4x4x(512 * 2 + (512 + 512)), 2, 1 & 64x64x512   \\ \hline
           & Attention & re deconv2 \& deconv2 & 64x64x512   \\ \hline
           & deconv3   & 4x4x(512 * 2 + (256 + 256)), 2, 1 & 128x128x256 \\ \hline
           & Attention & re deconv3 \& deconv3  & 128x128x256 \\ \hline
           & deconv4   & 4x4x(256 * 2 + (128 + 128)), 2, 1 & 256x256x128 \\ \hline
           & Attention & re deconv4 \& decovn4  & 256x256x128 \\ \hline
           & conv6     & 3x3x(128 * 2 + (64 + 64)), 1, 1  & 256x256x64  \\ \hline
           & conv6     & 3x3x1, 1, 1               & 256x256x1   \\ \hline
Initial Shading Output     & Attention & re output \& conv6    & 256x256x1   \\ \hline
\end{tabular}%
}
\label{tab:init_shading_def}
\end{table}

\paragraph{\textbf{Final Correction Module}:} This module consists of two encoder and decoder structures: i) The reflectance edge encoder ($\mathcal{F}_{R_E Enc}$), ii) the initial reflectance estimation encoder ($\mathcal{F}_{R_1}$), iii) the initial shading estimation encoder ($\mathcal{F}_{S_1}$), iv) the final reflectance decoder and v) the final shading decoder. The encoders use the previously introduced configurations detailed in table~\ref{tab:encoder_def}. Tables~\ref{tab:final_reflec_decoder_def} and~\ref{tab:final_shad_decoder_def} gives an overview of the final decoder configurations. Just like the previous module, a joint decoder structure is used. The reflectance receives skip connections from $\mathcal{F}_{R_1}$. In addition, the features from $\mathcal{F}_{R_1}$ and $\mathcal{F}_{R_E Enc}$ are passed through an attention layer and forwarded as an additional skip connection. Similarly, the shading decoder receives skip connections from $\mathcal{F}_{S_1}$. 

\begin{table}
\centering
\caption{Overview of the feature calibrator. It has a separate 1x1 convolutions for the reflectance}
\resizebox{0.8\textwidth}{!}{%
\begin{tabular}{|c|c|l|l|}
\hline
Name & \textbf{Layer} & \multicolumn{1}{c|}{\textbf{Kernel Size, Stride, Padding}} & \multicolumn{1}{c|}{\textbf{Output Size}} \\ \hline
\begin{tabular}[c]{@{}c@{}}Reflectance\\ Input\end{tabular}      & reflec conv1 & 1x1x(3 + 3), 1, 0 & 256x256x8  \\ \hline
\begin{tabular}[c]{@{}c@{}}Reflectance\\ Bottleneck\end{tabular} & reflec conv1 & 1x1x16, 1, 0      & 256x256x16 \\ \hline
\end{tabular}%
}
\label{tab:feature_calibrator}
\end{table}

The input to the reflectance encoder is the concatenation of the reflectance edge and the initial reflectance estimate from the previous decoder. This concatenated output is first fed through a feature calibration module (detailed in table~\ref{tab:feature_calibrator}) before being passed to the final reflectance decoder.

\begin{table}
\centering
\caption{Overview of the configuration for the final reflectance correction decoder}
\resizebox{0.8\textwidth}{!}{%
\begin{tabular}{|l|l|l|l|}
\hline
\textbf{Name} & \textbf{Layer} & \textbf{Kernel Size, Stride, Padding}                                    & \textbf{Output Size} \\ \hline
BottleNeck & deconv1   & 4x4x512, 2, 1                   & 32x32x512   \\ \hline
              & Attention      & \begin{tabular}[c]{@{}l@{}}$\mathcal{F}_{R_1}$ conv4\\ \&  $\mathcal{F}_{R_E Enc}$ conv4\end{tabular} & 32x32x512            \\ \hline
           & deconv2   & 4x4x(512 * 2 + (512 + 512)), 2, 1 & 64x64x512   \\ \hline
           & Attention & \begin{tabular}[c]{@{}l@{}}$\mathcal{F}_{R_1}$ conv3\\ \& $\mathcal{F}_{R_E Enc}$ conv3\end{tabular}              & 64x64x512   \\ \hline
           & deconv3   & 4x4x(512 * 2 + (256 + 256)), 2, 1 & 128x128x256 \\ \hline
           & Attention & \begin{tabular}[c]{@{}l@{}}$\mathcal{F}_{R_1}$ conv2\\ \& $\mathcal{F}_{R_E Enc}$ conv2\end{tabular}              & 128x128x256 \\ \hline
           & deconv4   & 4x4x(256 * 2 + (128 + 128)), 2, 1 & 256x256x128 \\ \hline
           & Attention & \begin{tabular}[c]{@{}l@{}}$\mathcal{F}_{R_1}$ conv1\\ \& $\mathcal{F}_{R_E Enc}$ conv1\end{tabular}              & 256x256x128 \\ \hline
           & conv6     & 3x3x(128 * 2 + (64 + 64)), 1, 1    & 256x256x64  \\ \hline
Final Reflectance Output           & conv6     & 3x3x3, 1, 1                           & 256x256x3   \\ \hline
\end{tabular}%
}
\label{tab:final_reflec_decoder_def}
\end{table}

\begin{table}
\centering
\caption{Overview of the configuration for the final shading correction decoder}
\resizebox{0.8\textwidth}{!}{%
\begin{tabular}{|l|l|l|l|}
\hline
\textbf{Name} & \textbf{Layer} & \textbf{Kernel Size, Stride, Padding}                                    & \textbf{Output Size} \\ \hline
BottleNeck & deconv1   & 4x4x512, 2, 1                   & 32x32x512   \\ \hline
              & Attention      & \begin{tabular}[c]{@{}l@{}}$\mathcal{F}_{R_1}$ conv4\\ \&  $\mathcal{F}_{R_E Enc}$ conv4\end{tabular} & 32x32x512            \\ \hline
           & deconv2   & 4x4x(512 * 2 + (512 + 512)), 2, 1 & 64x64x512   \\ \hline
           & Attention & \begin{tabular}[c]{@{}l@{}}$\mathcal{F}_{R_1}$ conv3\\ \& $\mathcal{F}_{R_E Enc}$ conv3\end{tabular}              & 64x64x512   \\ \hline
           & deconv3   & 4x4x(512 * 2 + (256 + 256)), 2, 1 & 128x128x256 \\ \hline
           & Attention & \begin{tabular}[c]{@{}l@{}}$\mathcal{F}_{R_1}$ conv2\\ \& $\mathcal{F}_{R_E Enc}$ conv2\end{tabular}              & 128x128x256 \\ \hline
           & deconv4   & 4x4x(256 * 2 + (128 + 128)), 2, 1 & 256x256x128 \\ \hline
           & Attention & \begin{tabular}[c]{@{}l@{}}$\mathcal{F}_{R_1}$ conv1\\ \& $\mathcal{F}_{R_E Enc}$ conv1\end{tabular}              & 256x256x128 \\ \hline
           & conv6     & 3x3x(128 * 2 + (64 + 64)), 1, 1    & 256x256x64  \\ \hline
Final Shading Output           & conv6     & 3x3x1, 1, 1                           & 256x256x1   \\ \hline
\end{tabular}%
}
\label{tab:final_shad_decoder_def}
\end{table}

\section{Supplementary: Loss Function Details}

The losses are broadly grouped into i) initial estimation loss ($\mathcal{L}_e$ \& $\mathcal{L}_i$), ii) final correction loss ($\mathcal{L}_f$), and iii) invariant and homogeneity constraint loss ($\mathcal{L}_{Norm}$ \& $\mathcal{L}_{TV}$). All losses use the standard MSE loss.

\paragraph{\textbf{1) Initial Estimation Loss}:} The initial estimation block outputs consist of the scales of reflectance edges, along with the initial reflectance and shading estimations. The reflectance edge outputs consist of $3$ outputs, $64\times64$, $128\times128$ and the full resolution of $256\times256$. The total edge loss is defined by:
\begin{equation}
    \label{eq:edge_loss}
    \mathcal{L}_e = \mathcal{L}_{RE_{64}} + \mathcal{L}_{RE_{128}} + \mathcal{L}_{RE_{256}}
\end{equation}

\noindent where $\mathcal{L}_{RE_{64}},~\mathcal{L}_{RE_{128}},~\mathcal{L}_{RE_{256}}$ are the losses on the reflectance edges with resolution of $64\times64$, $128\times128$ and $256\times256$. The ground truth for these losses is generated by using a Canny Edge operation on the reflectance ground truth. 
The initial estimations are matched to the ground truth reflectance and shading, hence:
\begin{equation}
    \label{eq:unrefined}
    \mathcal{L}_i = \mathcal{L}_{iR} + \mathcal{L}_{iS}
\end{equation}

\noindent where $\mathcal{L}_{iR}$ is the initial reflectance estimation and $\mathcal{L}_{iS}$ is the initial shading estimation losses.

\paragraph{\textbf{2) Final Correction Loss}:} For the final outputs of the network a reconstruction loss is added. This ensures that the network can learn the image formation model. On top of this loss, the reflectance and shading are also compared to the ground truth. The final loss is:
\begin{equation}
    \label{eq:refined}
    \mathcal{L}_f = \mathcal{L}_{R} + \mathcal{L}_{S} + \mathcal{L}_{rec}
\end{equation}

\noindent where $\mathcal{L}_{R}$ and $\mathcal{L}_{S}$ are the losses between the ground truth reflectance and shading and the network prediction. The $\mathcal{L}_{rec}$ loss is the reconstruction loss between the product of the network prediction and the input image.

\paragraph{\textbf{3) Invariant and Homogeneity Constraint Loss}:} In addition to the standard losses, additional constrain specific losses are also applied to explicitly encourage invariance and homogeneity in the reflectance. Given a segment, obtained from the segmentation map, the normalised $RGB$ values are compared between the prediction and the image. It is reasoned that since normalised $RGB$ are illumination invariant features, the predicted reflectance should in turn have similar values for the corresponding pixels:
\begin{equation}
    \label{eq:norm_rgb}
    \mathcal{L}_{Norm} = \sum^N_{s=1} \mathcal{L}_{nRGB}
\end{equation}

\noindent where, $\mathcal{L}_{nRGB}$ is the loss between the normalised $RGB$ of the predicted reflectance and the image, for the pixels in the segment belonging to class $s$.

In addition, large classes, especially indoors, like walls and ceilings, consists of largely homogeneous regions. Explicit homogeneous supervision is provided in the form of a Total Variation loss:

\begin{equation}
    \label{eq:tv}
    \mathcal{L}_{TV} = \sum^{N}_{c=ceil, wall}\mathcal{L}_{TV}(\hat{R}, R)
\end{equation}

\noindent where, $\mathcal{L}_{TV}$ is the MSE between the total variation of the predicted reflectance ($\hat{R}$) and the ground truth reflectance ($R$), for the pixels belonging to the ceiling and wall class.

Finally, to make the network produce perceptually consistent outputs, a perceptual loss with a pretrained VGG16~\cite{Simonyan2015} and Structural Dissimilarity loss is also added:

\begin{equation}
    \label{eq:dssim_percep}
    \mathcal{L}_{\delta} = \lambda_{P} \; \mathcal{L}_{P} + \lambda_{dssim} \; \mathcal{L}_{dssim}
\end{equation}

\noindent where $\mathcal{L}_{P}$ is the perceptual loss between the predicted reflectance and the ground truth reflectance and $\mathcal{L}_{dssim}$ is the Structural dissimilarity metric between the predicted and ground truth reflectance and shading. $\lambda_{P}$ and $\lambda_{dssim}$ are empirically set as $0.05$ and $0.4$, respectively.

The final loss to minimise for the network then becomes:
\begin{equation}
\begin{aligned}
    \label{eq:total_supp}
    \mathcal{L} = \lambda_e \; \mathcal{L}_e + \lambda_i \; \mathcal{L}_i + \mathcal{L}_f \\
    + \mathcal{L}_{Norm} + \mathcal{L}_{TV} + \mathcal{L}_{\delta}
\end{aligned}
\end{equation}

\noindent where $\lambda_e$ and $\lambda_i$ are empirically found to be optimum as $0.4$ and $0.5$, respectively.

\section{Supplementary: Additional Visuals}

\subsection{IIW}

\begin{figure}
    \centering
    \includegraphics[width=\linewidth]{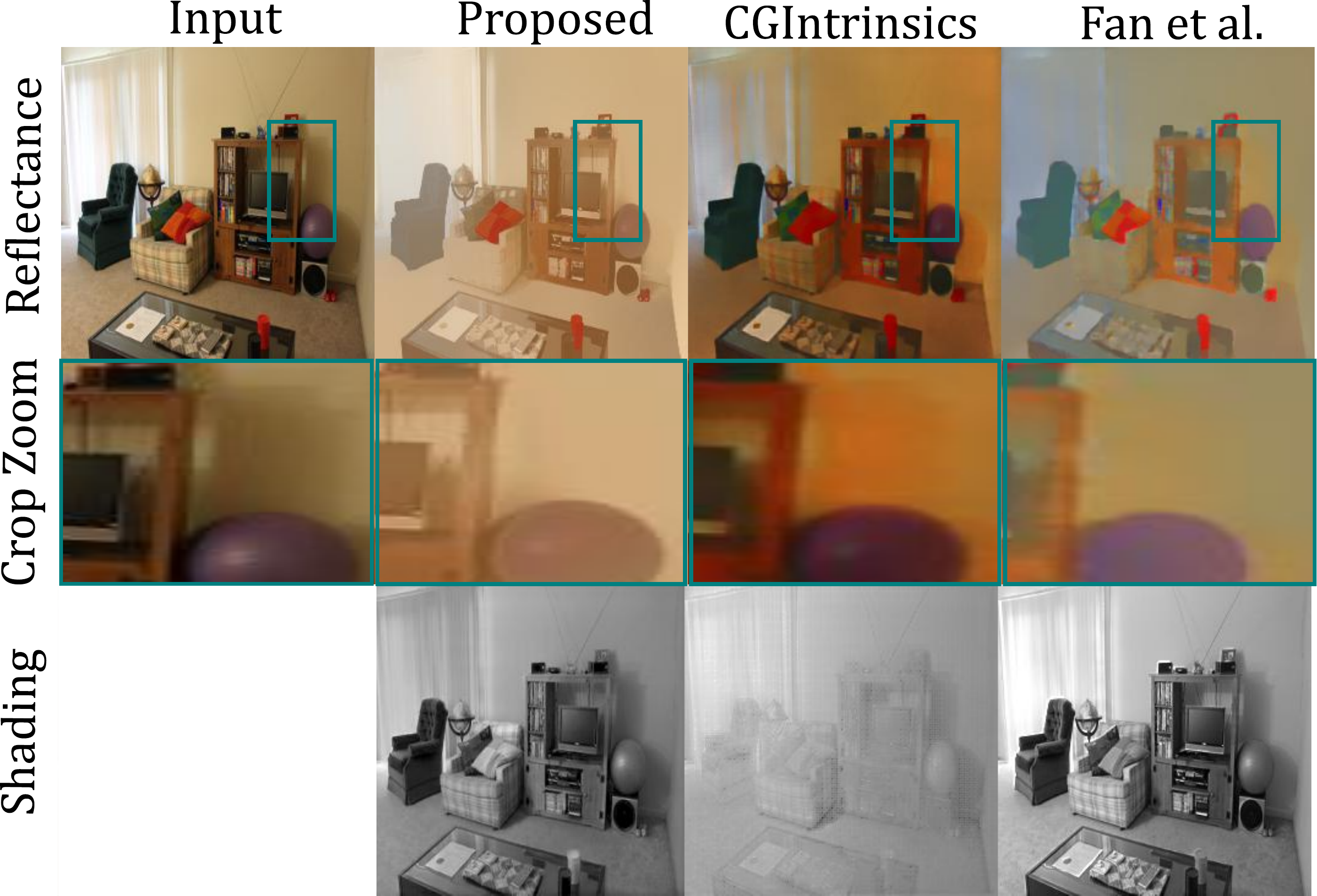}
    \caption{The shadow behind the TV case is completely removed by the proposed method, while the competing method suffers from artefacts.}
    \label{fig:iiw_supp5}
\end{figure}

\begin{figure}
    \centering
    \includegraphics[width=\linewidth]{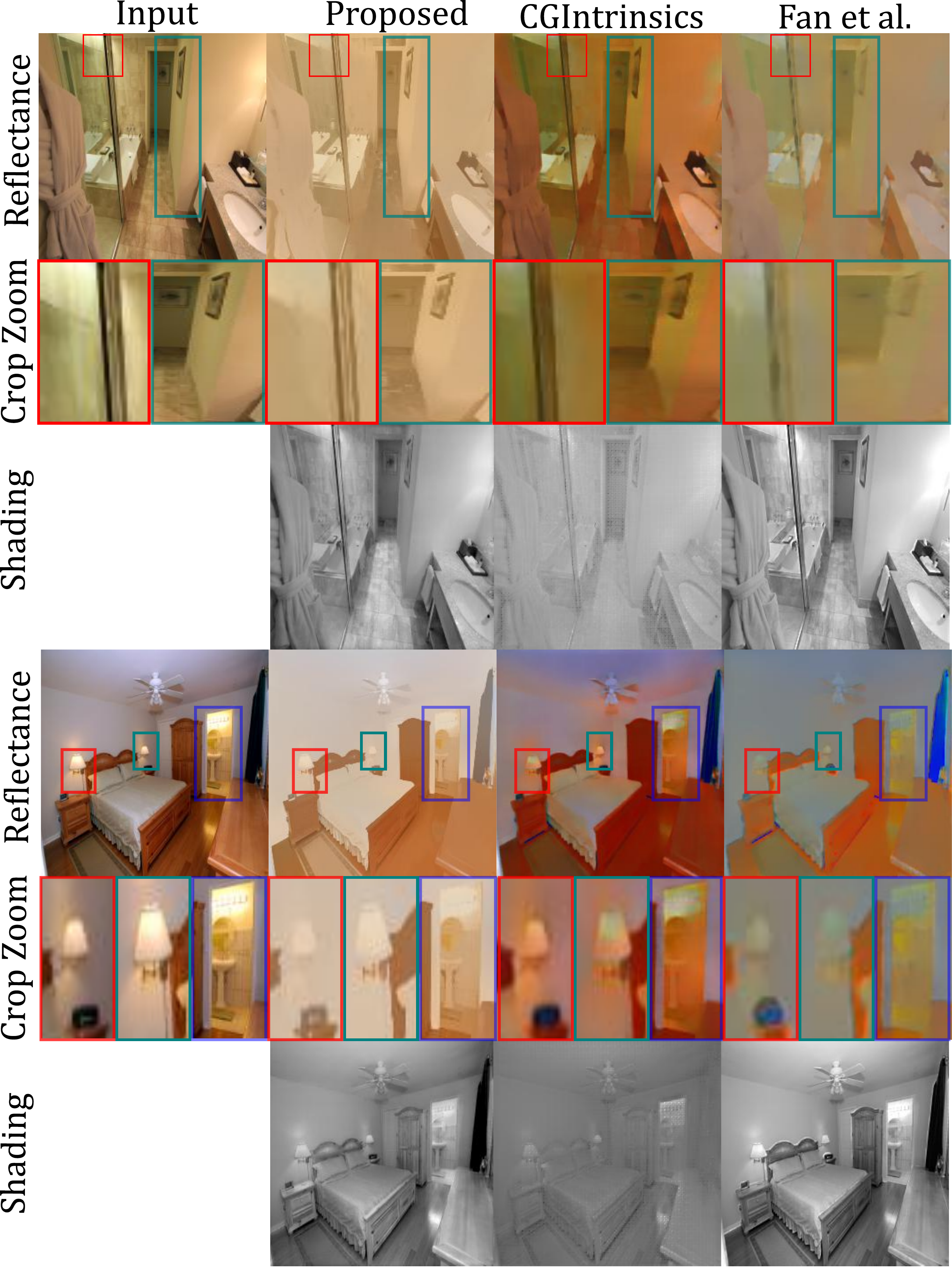}
    \caption{Additional visuals from the IIW test-set. The second row for each image group shows the zoom of the highlighted box. The proposed method can recover reflectance from various illumination effects including coloured illumination.}
    \label{fig:iiw_supp1}
\end{figure}

\begin{figure}
    \centering
    \includegraphics[width=\linewidth]{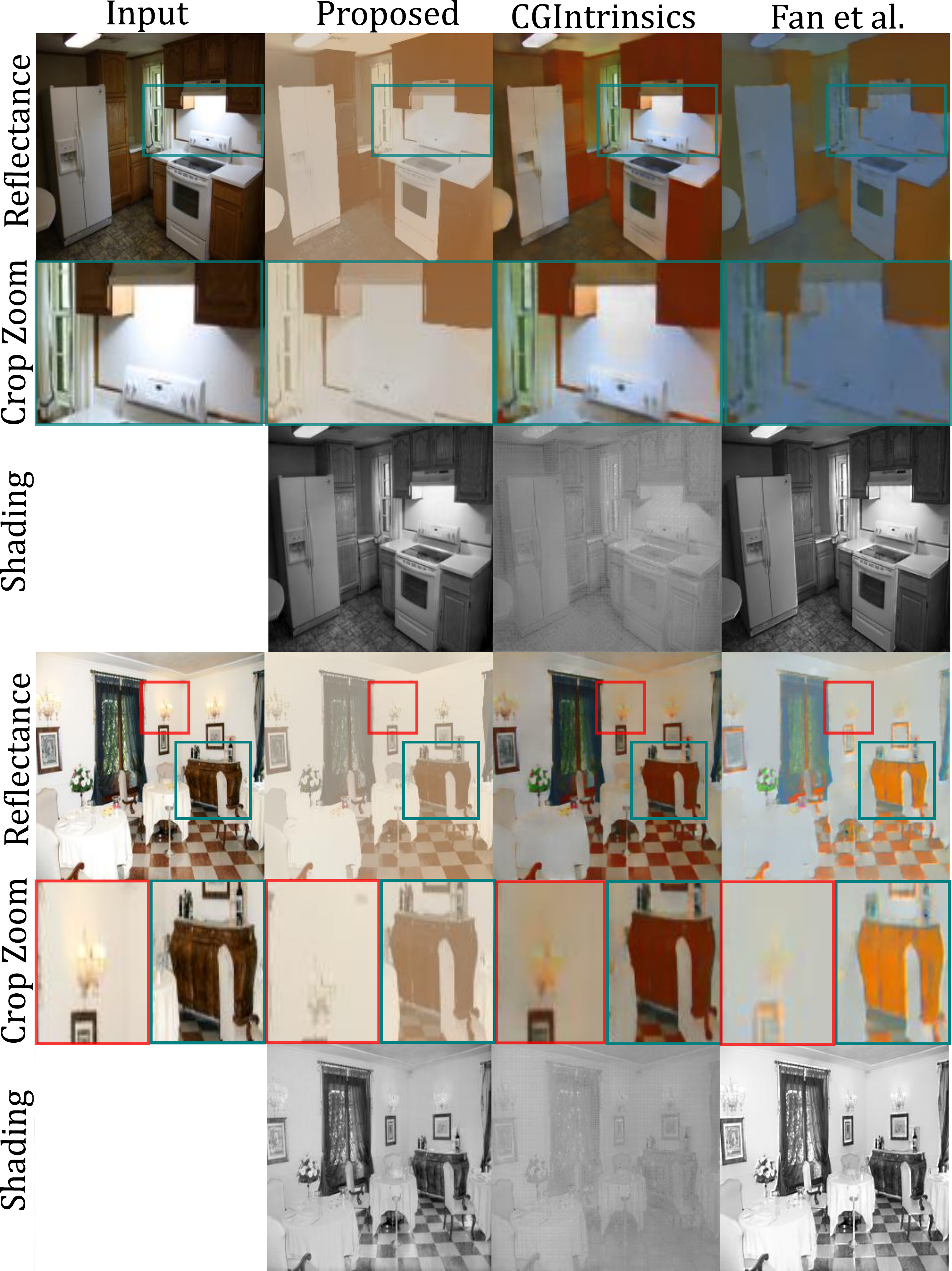}
    \caption{The proposed method can preserve reflectance colour while also being able to handle various illumination effects.}
    \label{fig:iiw_supp2}
\end{figure}

\begin{figure}
    \centering
    \includegraphics[width=\linewidth]{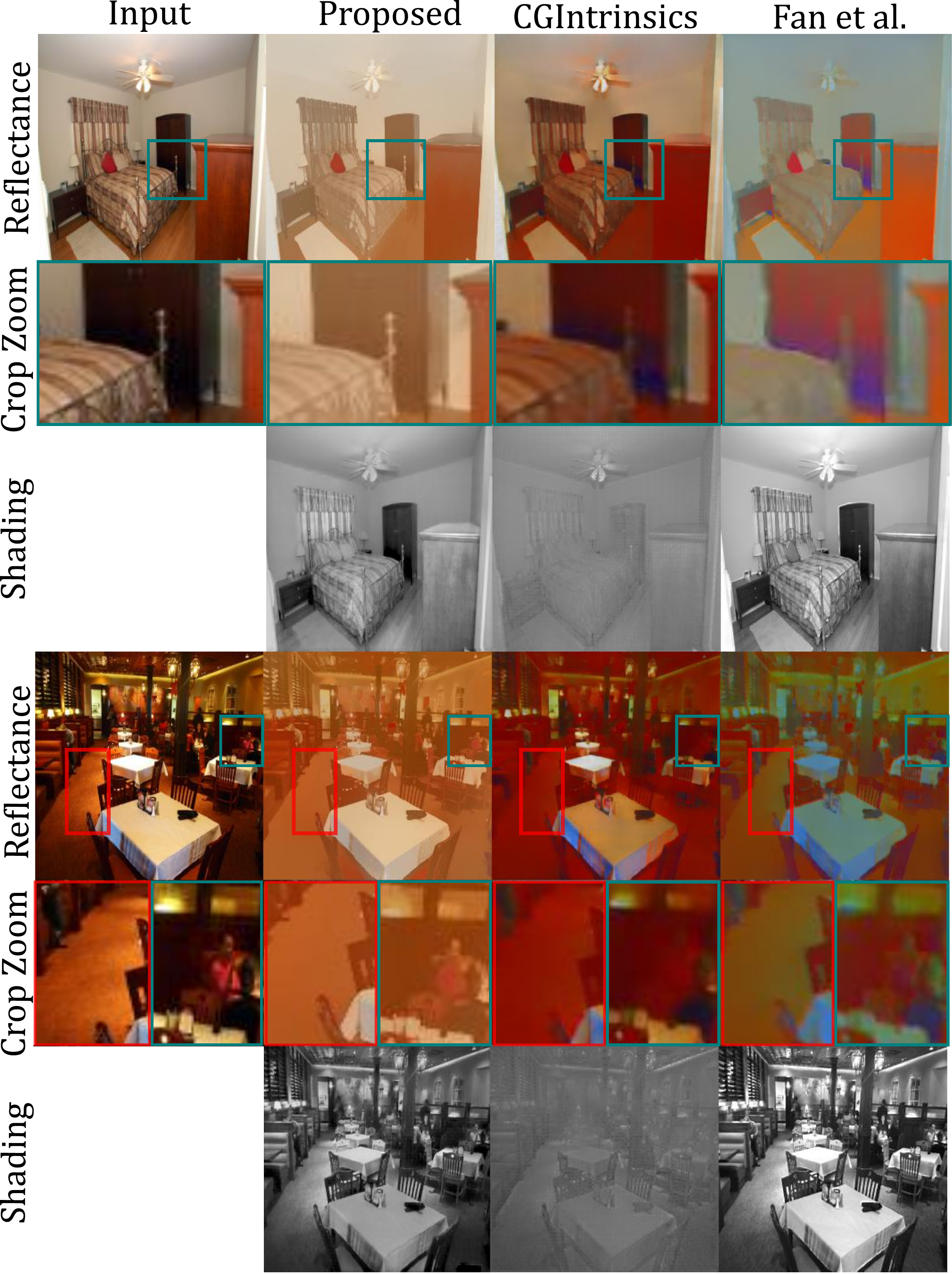}
    \caption{The proposed method's reflectance is free from artefacts, while the competing method has that problem. Additionally, background details are preserved in sharper clarity, compared to the competing methods.}
    \label{fig:iiw_supp3}
\end{figure}

\begin{figure}
    \centering
    \includegraphics[width=\linewidth]{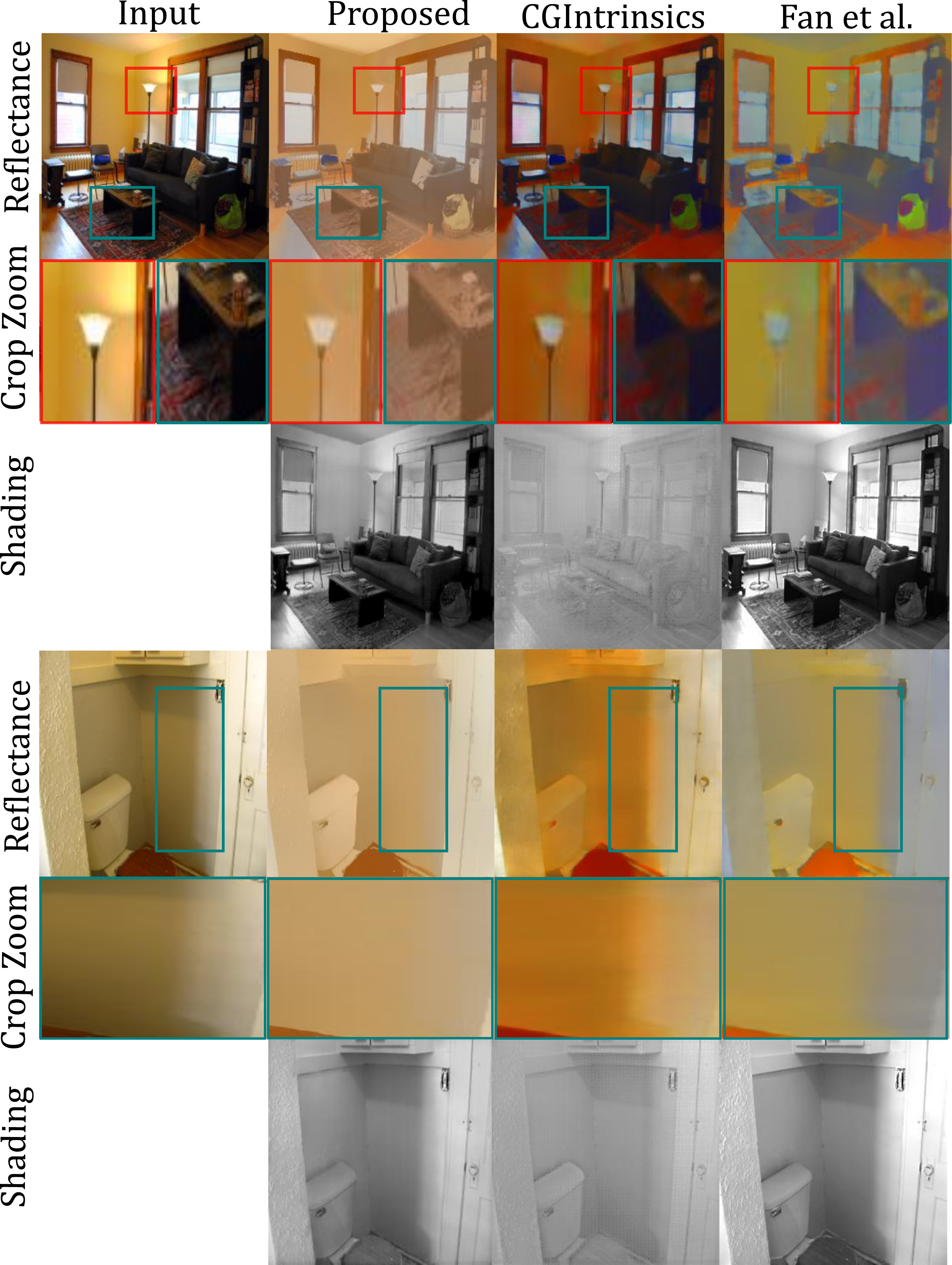}
    \caption{The proposed method can disentangle the shading and illumination effects, giving a comparatively flatter and piece-wise constant reflectance.}
    \label{fig:iiw_supp4}
\end{figure}

Additional visual comparisons with the baselines, for the IIW test-set, are provided in figs~\ref{fig:iiw_supp5}, \ref{fig:iiw_supp2}, \ref{fig:iiw_supp3}, \ref{fig:iiw_supp4} and~\ref{fig:iiw_supp1}.

\subsection{Dataset}

\begin{figure}
    \centering
    \includegraphics[width=\linewidth]{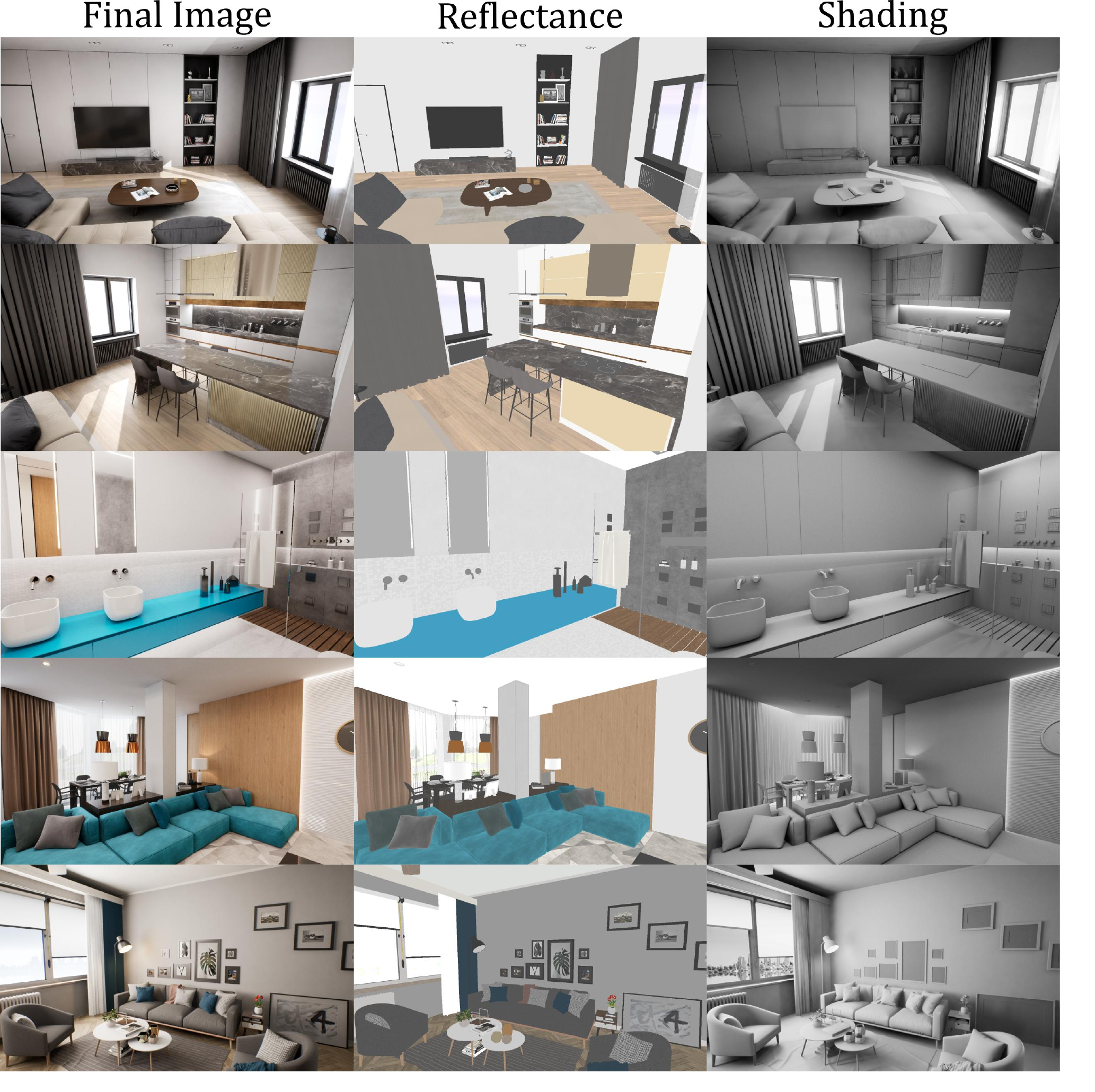}
    \caption{Visuals from the proposed dataset. For each of the images, the dense ground truth reflectance and the shading is also provided.}
    \label{fig:dataset_supp1}
\end{figure}

\begin{figure}
    \centering
    \includegraphics[width=\linewidth]{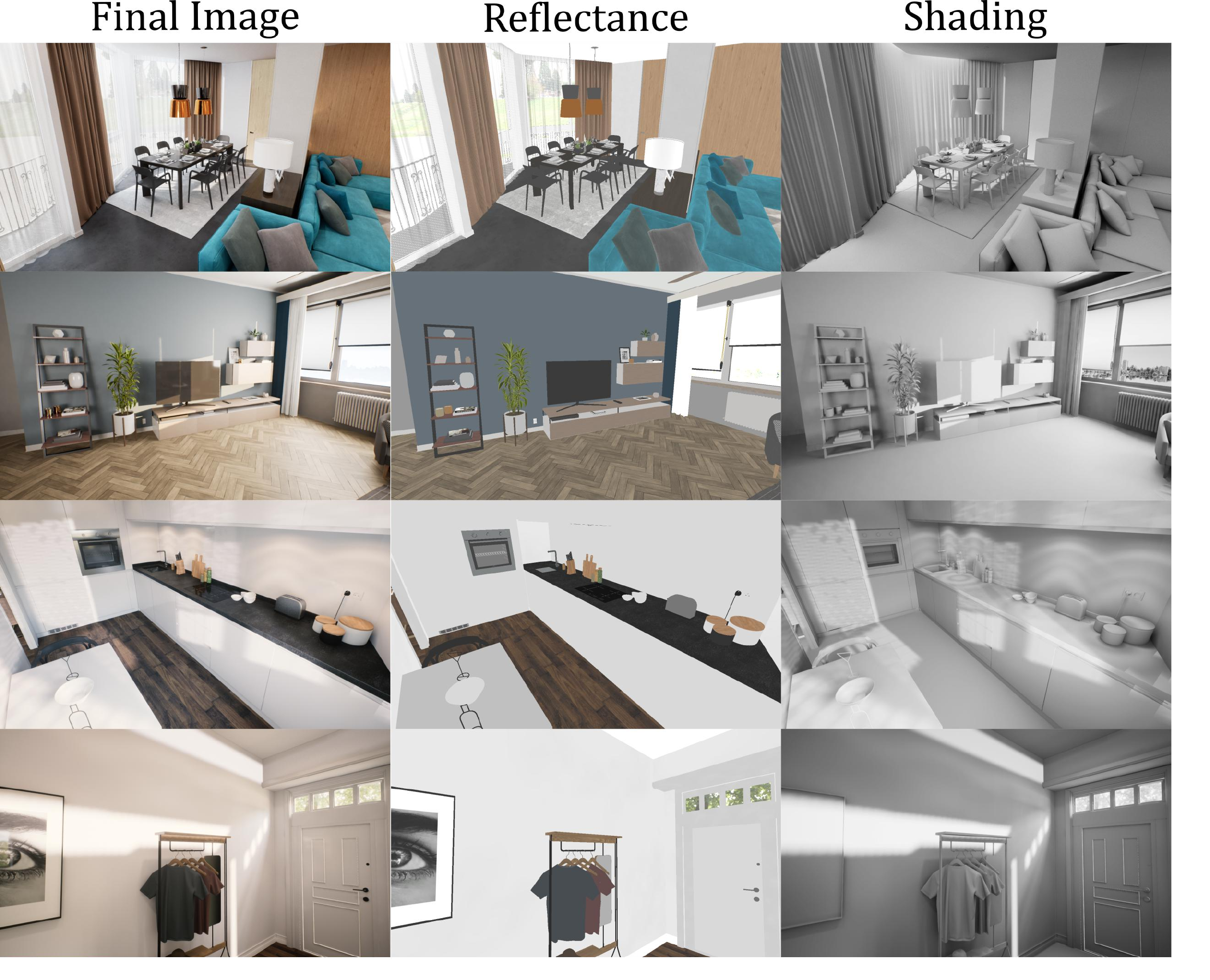}
    \caption{Additional visuals from the dataset. The images are of realistic indoor scenes, consisting of various settings like living rooms, kitchen, and hallways.}
    \label{fig:dataset_supp2}
\end{figure}

\begin{figure}
    \centering
    \includegraphics[width=\linewidth]{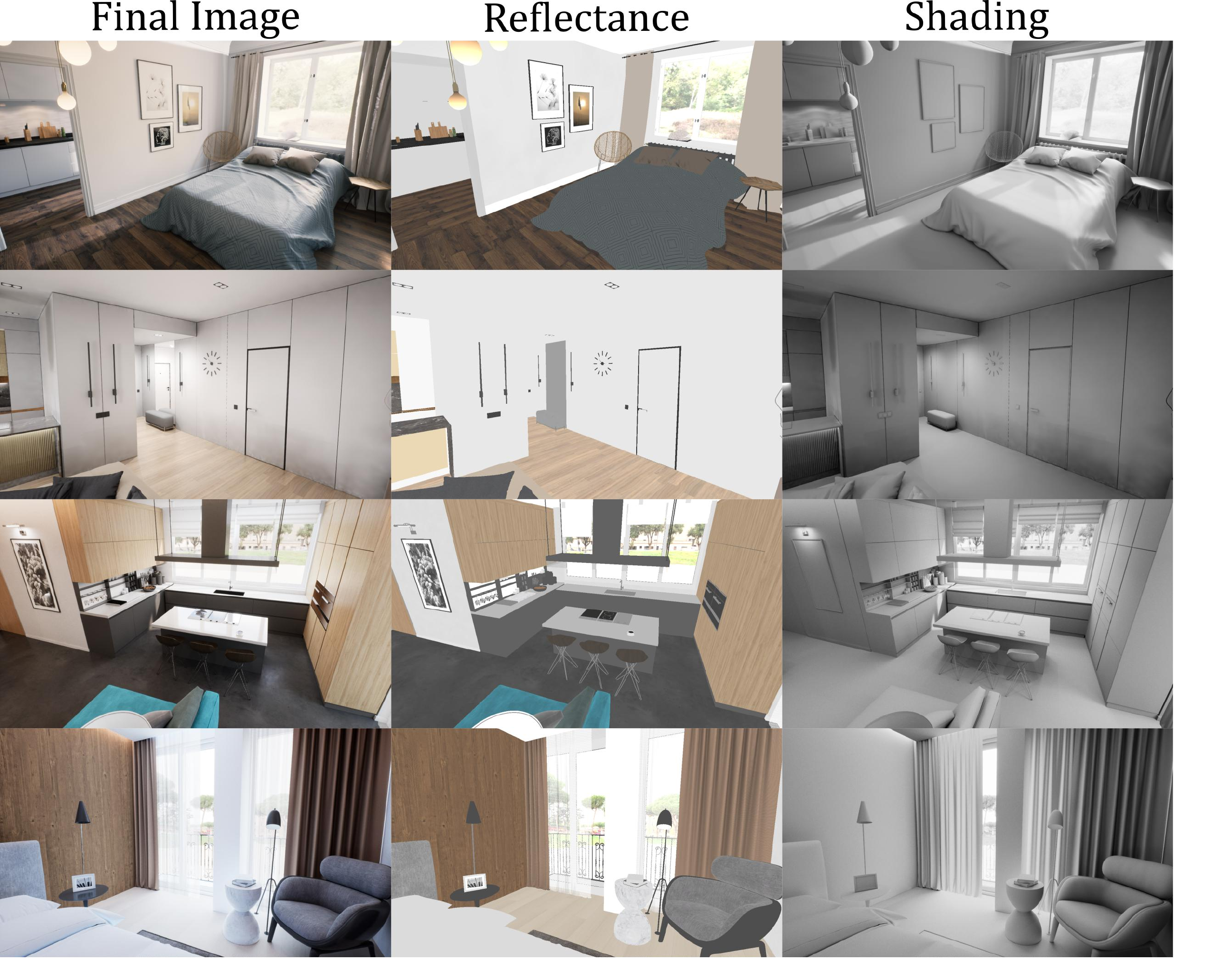}
    \caption{More visuals from the dataset showing different bedroom and kitchen styles.}
    \label{fig:dataset_supp3}
\end{figure}

Figs~\ref{fig:dataset_supp1}, \ref{fig:dataset_supp2} and~\ref{fig:dataset_supp3} provides some additional visuals for the proposed dataset, along with the corresponding reflectance and shading ground truth that is made available with this dataset.

\clearpage
%
%
\bibliographystyle{splncs04}
\bibliography{egbib}

\begin{thebibliography}{10}
\providecommand{\url}[1]{\texttt{#1}}
\providecommand{\urlprefix}{URL }
\providecommand{\doi}[1]{https://doi.org/#1}

\bibitem{Barron2015}
Barron, J.T., Malik, J.: Shape, illumination, and reflectance from shading.
  IEEE TPAMI pp. 1670--1687 (2015)

\bibitem{Baslamisli2018ECCV}
Baslamisli, A.S., Groenestege, T.T., Das, P., Le, H.A., Karaoglu, S., Gevers,
  T.: Joint learning of intrinsic images and semantic segmentation. In: ECCV
  (2018)

\bibitem{Baslamisli2019}
Baslamisli, A.S., Das, P., Le, H., Karaoglu, S., Gevers, T.: Shadingnet: Image
  intrinsics by fine-grained shading decomposition. IJCV  \textbf{129},
  2445--2473 (2021)

\bibitem{Baslamisli2020}
Baslamisli, A.S., Liu, Y., Karaoglu, S., Gevers, T.: Physics-based shading
  reconstruction for intrinsic image decomposition. Comput. Vis. and Image
  Understanding pp. 1--14 (2020)

\bibitem{Beigpour2011ObjectRB}
Beigpour, S., van~de Weijer, J.: Object recoloring based on intrinsic image
  estimation. ICCV pp. 327--334 (2011)

\bibitem{Bell2014}
Bell, S., Bala, K., Snavely, N.: Intrinsic images in the wild. ACM TOG  (2014)

\bibitem{Bi2015}
Bi, S., Han, X., Yu, Y.: An l1 image transform for edge-preserving smoothing
  and scene-level intrinsic decomposition. ACM TOG  \textbf{34}(4) (Jul 2015).
  \doi{10.1145/2766946}, \url{https://doi.org/10.1145/2766946}

\bibitem{Bonneel2014}
Bonneel, N., Sunkavalli, K., Tompkin, J., Sun, D., Paris, S., Pfister, H.:
  Interactive intrinsic video editing. ACM TOG pp. 197:1--197:10 (2014)

\bibitem{cheng2021}
Cheng, B., Misra, I., Schwing, A.G., Kirillov, A., Girdhar, R.:
  Masked-attention mask transformer for universal image segmentation. arXiv
  (2021)

\bibitem{Cheng2019}
Cheng, Z., Zheng, Y., You, S., Sato, I.: Non-local intrinsic decomposition with
  near-infrared priors. In: ICCV (October 2019)

\bibitem{unrealengine}
{Epic Games}: Unreal engine. -  (-), \url{https://www.unrealengine.com}

\bibitem{Fan2018}
Fan, Q., Yang, J., Hua, G., Chen, B., Wipf, D.: Revisiting deep intrinsic image
  decompositions. In: CVPR (2018)

\bibitem{Finlayson1992}
Finlayson, G.D.: Colour Object Recognition. Master's thesis, Simon Fraser
  University (1992)

\bibitem{Garces2012}
Garces, E., Munoz, A., Lopez-Moreno, J., Gutierrez, D.: Intrinsic images by
  clustering. Comput. Graph. Forum (Proceedings of the Eurographics Symposium
  on Rendering)  \textbf{31}(4) (2012),
  \url{http://www-sop.inria.fr/reves/Basilic/2012/GMLG12}

\bibitem{Gehler2011}
Gehler, P.V., Rother, C., Kiefel, M., Zhang, L., Schölkopf, B.: Recovering
  intrinsic images with a global sparsity prior on reflectance. In: NeurIPS
  (2011)

\bibitem{Gevers1999}
Gevers, T., Smeulders, A.: Color-based object recognition. PR pp. 453--464
  (1999)

\bibitem{Grosse2009}
Grosse, R., Johnson, M.K., Adelson, E.H., Freeman, W.T.: Ground truth dataset
  and baseline evaluations for intrinsic image algorithms. In: ICCV (2009)

\bibitem{henderson2019}
Henderson, P., Ferrari, V.: Learning single-image 3d reconstruction by
  generative modelling of shape, pose and shading. International Journal of
  Computer Vision  (2019)

\bibitem{Jeon2014}
Jeon, J., Cho, S., Tong, X., Lee, S.: Intrinsic image decomposition using
  structure-texture separation and surface normals. In: ECCV (2014)

\bibitem{kingma2014}
Kingma, D.P., Ba, J.: Adam: A method for stochastic optimization (2014),
  \url{http://arxiv.org/abs/1412.6980}, cite arxiv:1412.6980Comment: Published
  as a conference paper at the 3rd International Conference for Learning
  Representations, San Diego, 2015

\bibitem{Land1971}
Land, E.H., McCann, J.J.: Lightness and retinex theory. J. of Optical Society
  of America pp. 1--11 (1971)

\bibitem{Lee2012}
Lee, K.J., Zhao, Q., Tong, X., Gong, M., Izadi, S., Lee, S.U., Tan, P., Lin,
  S.: Estimation of intrinsic image sequences from image+depth video. In:
  Fitzgibbon, A., Lazebnik, S., Perona, P., Sato, Y., Schmid, C. (eds.) ECCV.
  pp. 327--340. Springer Berlin Heidelberg, Berlin, Heidelberg (2012)

\bibitem{Li2018ECCV}
Li, Z., Snavely, N.: Cgintrinsics: Better intrinsic image decomposition through
  physically-based rendering. In: ECCV (2018)

\bibitem{Li2020}
Li, Z., Shafiei, M., Ramamoorthi, R., Sunkavalli, K., Chandraker, M.: Inverse
  rendering for complex indoor scenes: Shape, spatially-varying lighting and
  svbrdf from a single image. CVPR pp. 2472--2481 (2020)

\bibitem{Li2021CVPR}
Li, Z., Yu, T., Sang, S., Wang, S., Bi, S., Xu, Z., Yu, H., Sunkavalli, K.,
  Hasan, M., Ramamoorthi, R., Chandraker, M.: Openrooms: An end-to-end open
  framework for photorealistic indoor scene datasets. CoRR
  \textbf{abs/2007.12868} (2020), \url{https://arxiv.org/abs/2007.12868}

\bibitem{Narihia2015}
Narihira, T., Maire, M., Yu, S.X.: Direct intrinsics: Learning albedo-shading
  decomposition by convolutional regression. In: ICCV (2015)

\bibitem{Narihira2015-2}
{Narihira}, T., {Maire}, M., {Yu}, S.X.: Learning lightness from human
  judgement on relative reflectance. In: CVPR. pp. 2965--2973 (June 2015).
  \doi{10.1109/CVPR.2015.7298915}

\bibitem{Nestmeyer2016}
Nestmeyer, T., Gehler, P.V.: Reflectance adaptive filtering improves intrinsic
  image estimation. CoRR  \textbf{abs/1612.05062} (2016),
  \url{http://arxiv.org/abs/1612.05062}

\bibitem{roberts2021}
Roberts, M., Ramapuram, J., Ranjan, A., Kumar, A., Bautista, M.A., Paczan, N.,
  Webb, R., Susskind, J.M.: {Hypersim}: {A} photorealistic synthetic dataset
  for holistic indoor scene understanding. In: International Conference on
  Computer Vision (ICCV) 2021 (2021)

\bibitem{Saini2019}
Saini, S., Narayanan, P.J.: Semantic hierarchical priors for intrinsic image
  decomposition. CoRR  \textbf{abs/1902.03830} (2019),
  \url{http://arxiv.org/abs/1902.03830}

\bibitem{Sengupta2019}
Sengupta, S., Gu, J., Kim, K., Liu, G., Jacobs, D.W., Kautz, J.: Neural inverse
  rendering of an indoor scene from a single image. CoRR
  \textbf{abs/1901.02453} (2019), \url{http://arxiv.org/abs/1901.02453}

\bibitem{Shafer1985}
Shafer, S.: Using color to separate reflection components. Color Research and
  App. pp. 210--218 (1985)

\bibitem{Shi2017}
Shi, J., Dong, Y., Su, H., Yu, S.X.: Learning non-lambertian object intrinsics
  across shapenet categories. In: CVPR (2017)

\bibitem{shu2017}
Shu, Z., Yumer, E., Hadap, S., Sunkavalli, K., Shechtman, E., Samaras, D.:
  Neural face editing with intrinsic image disentangling. CoRR
  \textbf{abs/1704.04131} (2017), \url{http://arxiv.org/abs/1704.04131}

\bibitem{Simonyan2015}
Simonyan, K., Zisserman, A.: Very deep convolutional networks for large-scale
  image recognition. In: ICLR (2015)

\bibitem{tang2020}
Tang, H., Qi, X., Xu, D., Torr, P.H.S., Sebe, N.: Edge guided gans with
  semantic preserving for semantic image synthesis. CoRR  (2020)

\bibitem{Xie2015}
"Xie, S., Tu, Z.: Holistically-nested edge detection. In: ICCV (2015)

\bibitem{Xu2020}
Xu, J., Hou, Y., Ren, D., Liu, L., Zhu, F., Yu, M., Wang, H., Shao, L.: Star: A
  structure and texture aware retinex model. IEEE TIP pp. 5022--5037 (2020)

\bibitem{Zhao2012}
{Zhao}, Q., {Tan}, P., {Dai}, Q., {Shen}, L., {Wu}, E., {Lin}, S.: A
  closed-form solution to retinex with nonlocal texture constraints. IEEE TPAMI
   \textbf{34}(7),  1437--1444 (July 2012). \doi{10.1109/TPAMI.2012.77}

\bibitem{Zhou2019}
Zhou, H., Yu, X., Jacobs, D.W.: Glosh: Global-local spherical harmonics for
  intrinsic image decomposition. In: ICCV (October 2019)

\end{thebibliography}
\end{document}